\documentclass[]{youtu} % For LaTeX2e

\usepackage{mathpazo}
\usepackage{graphicx}
\usepackage[numbers]{natbib}
\usepackage{algorithm}
\usepackage{algorithmic}
\newtheorem{proposition}{Proposition}

\usepackage[table,dvipsnames]{xcolor}
\usepackage[most]{tcolorbox}
\tcbuselibrary{listings, breakable}
\usepackage{booktabs,multirow,subcaption}
\usepackage{multirow}
\usepackage{amssymb}

\usepackage{hyperref}
\usepackage{url}

\lstdefinestyle{pythonstyle}{
    language=Python,
    basicstyle=\ttfamily\small,
    numbers=none,
    numberstyle=\tiny\color{gray},
    keywordstyle=\color{blue!70!black},
    commentstyle=\color{green!50!black},
    stringstyle=\color{orange!80!black},
    breaklines=true,
    tabsize=2,
    showstringspaces=false,
}

\newtcblisting{pycodebox}{
  listing engine=listings,
  colback=blue!5!white,             % 淡蓝背景
  colframe=MidnightBlue!85!black,  % 深蓝边框
  coltitle=white,                  
  colbacktitle=MidnightBlue!85!black,
  title=\textbf{Python Code},
  fonttitle=\bfseries,
  listing only,
  breakable,
  listing options={style=pythonstyle},
}

\newtcolorbox{outputbox}{
  colback=gray!10!white,       % 浅灰背景
  colframe=MidnightBlue!80!black, % 深蓝边框
  fonttitle=\bfseries,
  coltitle=white,
  colbacktitle=MidnightBlue!80!black,
  title=Output,
  left=4mm, right=4mm, top=2mm, bottom=2mm,
  boxrule=0.8pt,
  arc=3mm, % 圆角
}

\definecolor{hlblue}{RGB}{182,210,232} % 近似图中的浅蓝
\newcommand{\best}[1]{\cellcolor{hlblue}{#1}}

\title{Count Counts: Motivating Exploration in LLM Reasoning with Count-based Intrinsic Rewards}

\author{Xuan Zhang$^{1,2,3}$, Ruixiao Li$^{1,2}$, Zhijian Zhou$^{1,2,3}$, Long Li$^{1}$, Yulei Qin$^{3}$, Ke Li$^{3}$, Xing Sun$^{3}$, Xiaoyu Tan$^{3\dagger}$, Chao Qu$^{1\dagger}$, Yuan Qi$^{1,2\dagger}$} 

\affiliation{$^{1}$Fudan University, $^{2}$Shanghai Innovation Institute, $^{3}$Tencent Youtu Lab}

\date{October 23, 2025}
\correspondence{xuanzhang24@m.fudan.edu.cn}
\sourcecode{https://github.com/dd88s87/MERCI}
\project{https://dd88s87.github.io/MERCI.github.io}
\begin{document}

\abstract{Reinforcement Learning (RL) has become a compelling way to strengthen the multi step reasoning ability of Large Language Models (LLMs). However, prevalent RL paradigms still lean on sparse outcome-based rewards and limited exploration, which often drives LLMs toward repetitive and suboptimal reasoning patterns. In this paper, we study the central question of how to design exploration for LLM reasoning and introduce MERCI (\textbf{M}otivating \textbf{E}xploration in LLM \textbf{R}easoning with \textbf{C}ount-based \textbf{I}ntrinsic Rewards), a novel RL algorithm that augments policy optimization with a principled intrinsic reward. Building on the idea of count-based exploration, MERCI leverages a lightweight Coin Flipping Network (CFN) to estimate the pseudo count and further epistemic uncertainty over reasoning trajectories, and converts them into an intrinsic reward that values novelty while preserving the learning signal from task rewards. We integrate MERCI into some advanced RL frameworks like Group Relative Policy Optimization (GRPO). Experiments on complex reasoning benchmarks demonstrate that MERCI encourages richer and more varied chains of thought, significantly improves performance over strong baselines, and helps the policy escape local routines to discover better solutions. It indicates that our targeted intrinsic motivation can make exploration reliable for language model reasoning.}
\maketitle

% \dateandcorrespondence

\section{Introduction}

Reinforcement learning (RL) \citep{Sutton1998} has become a cornerstone of advancing the multi-step reasoning capabilities of Large Language Models (LLMs), enabling them to tackle complex domains like competitive mathematics and code generation \citep{jaech2024openai,guo2025deepseek,aime}. However, these tasks feature sparse rewards, with feedback available only after completing a lengthy reasoning chain, making exploration a critical challenge. Recent breakthroughs, such as Group Relative Policy Optimization (GRPO) \citep{deepseek-math} and Dynamic sAmpling Policy Optimization (DAPO) \citep{yu2025dapo}, have streamlined the training process by eliminating the need for an explicit value function. This yields local variability at the token level, but it does not produce exploration that is coherent across the length of a reasoning trajectory. To guide exploration in such frameworks, many prevalent techniques rely on entropy regularization to encourage local policy diversity. While effective, this approach is limited for complex, long-horizon tasks. We see an opportunity to design complementary strategies that provide more directed, temporally-consistent exploration signals particularly for those tasks, motivating our investigation into principled exploration strategies compatible with modern value-free RL.

The exploration-exploitation trade-off is a classic challenge in RL \citep{jin2018q,azar2017minimax}. Simple approaches such as $\epsilon$-greedy \citep{mnih2015human} or Boltzmann exploration with entropy-based regularization \citep{mnih2016asynchronous}, inject undirected noise to encourage stochasticity \citep{osband2016deep}. While these \emph{``shallow''} exploration methods visit all states theoretically, they can be exponentially inefficient in simple yet illustrative examples \citep{osband2016generalization,kakade2003sample}. In notoriously difficult exploration tasks like the video game \emph{Montezuma’s Revenge}, these methods fail because the chance of discovering long, precise action sequences needed for reward is vanishingly small. In contrast, \emph{``deep exploration''} strategies are both theoretically and empirically superior in such scenarios. These methods follow the principle of ``optimism in the face of uncertainty'' \citep{kearns2002near,brafman2002r,o2018uncertainty}, encouraging the agent to explore regions of the state-action space where its knowledge is limited. This is often implemented by generating an \textbf{intrinsic reward} to densify the sparse signal from the environment. Canonical examples include pseudo-counts \citep{bellemare2016unifying,ostrovski2017count}, Bootstrapped DQN \citep{osband2016deep}, Random Network Distillation (RND) \citep{burda2018exploration}, the intrinsic curiosity module (ICM) \citep{pathak2017curiosity}, and methods based on the Uncertainty Bellman Equation (UBE) \citep{o2018uncertainty}.

Although desirable, existing methods for estimating epistemic uncertainty \citep{mannor2007bias} do not scale to modern LLMs. Deep Ensembles \citep{osband2016deep,lakshminarayanan2017simple}, which train multiple independent models, are prohibitively expensive. Monte Carlo dropout \citep{gal2016dropout}, though cheaper, still adds significant inference overhead. Other methods face architectural or theoretical hurdles: pseudo-count techniques \citep{bellemare2016unifying,ostrovski2017count} depend on normalized probability densities and preclude efficient batching, while curiosity-driven methods \citep{burda2018exploration,pathak2017curiosity} lack theoretical guarantees on how the exploration bonus should decay. The UBE framework \citep{o2018uncertainty}, while principled, relies on estimating local uncertainty, a notoriously difficult task often relegated to heuristics. This fundamental mismatch between classic uncertainty quantification and the scale of LLMs necessitates a novel approach.

Our work is founded on a critical insight applicable to a broad class of LLM reasoning tasks—specifically those that are self-contained, such as mathematical problem-solving, where the model operates without an external, stochastic world. \textbf{In this context of autoregressive generation, the underlying Markov Decision Process (MDP) has known and deterministic transitions.} When an LLM in a state $s$ (the token sequence generated so far) selects an action $a$ (the next token), the subsequent state $s'=(s, a)$ is determined without ambiguity. This property dramatically simplifies the Uncertainty Bellman Equation, which propagates uncertainty from two sources: the reward function estimate ($\hat{r}$) and the transition function estimate ($\hat{P}$). With known transitions, the epistemic uncertainty of $\hat{P}$ is zero. The UBE thus reduces to a simple accumulation of local reward uncertainty along a trajectory. This reframes the intractable problem of estimating Q-value uncertainty into the more manageable one of estimating local reward uncertainty. To make this tangible, we propose to proxy this uncertainty using a measure of state novelty—a practical and effective approach in sparse-reward settings. To this end, we employ the ``Flipping Coins'' method \citep{Lobel2023Flipping}, a computationally lightweight and theoretically grounded pseudo-counting technique that provides a scalable estimator for this purpose. We formalize this entire approach in our proposed algorithm, MERCI (\textbf{M}otivating \textbf{E}xploration in LLM \textbf{R}easoning with \textbf{C}ount-based \textbf{I}ntrinsic Rewards).

To our knowledge, this is the first work to derive and apply a deep exploration algorithm for LLM reasoning directly from a principled simplification of the UBE. By recognizing that \textbf{the LLM serves as its own perfectly known world model}, we bridge the gap between model-aware RL theory and the typically model-free application of RL to LLMs. Our method integrates this simplified UBE framework with the ``Flipping Coins'' pseudo-count module to generate an \textbf{intrinsic reward}. This reward, expressed as an exploration bonus, guides policy optimization algorithms like GRPO to explore novel reasoning trajectories based on a coherent, temporally-consistent signal of epistemic uncertainty.  Experiments on complex reasoning benchmarks demonstrate that this approach significantly improves performance, effectively mitigating the tendency of standard algorithms to converge on repetitive and suboptimal solutions. 

Our main contributions are summarized as follows:
\begin{itemize}
    \item \textbf{A Novel Theoretical Framework for LLM Exploration.} We establish a new framework based on a key insight: the LLM's known and deterministic transition dynamics simplify the Uncertainty Bellman Equation. This renders principled, uncertainty-driven exploration tractable at scale by reducing the intractable problem of Q-value uncertainty to a manageable estimation of local reward uncertainty.
    
    \item \textbf{A Practical and Scalable Exploration Algorithm.} We propose \textbf{MERCI}, a novel algorithm that operationalizes our theoretical framework. MERCI employs a highly scalable counting method to translate state novelty into a potent intrinsic reward signal, designed for seamless integration with modern, value-free policy optimization methods like GRPO.
    
    \item \textbf{State-of-the-Art Performance on Complex Reasoning.} Our extensive empirical evidence on challenging reasoning benchmarks, including MATH and SQL generation, demonstrate that MERCI beats strong baselines. Its directed exploration mechanism mitigates premature convergence and leads to the discovery of more robust and accurate solutions.
\end{itemize}

\section{Preliminaries}
\paragraph{Coin Flip Network (CFN)}
\label{cfn}
CFN is a computationally efficient method of count-based exploration, which estimates a state's visitation count by solving a simple regression problem. The core idea is that a state's visitation count can be estimated by leveraging the statistical properties of the Rademacher distribution (i.e., random coin flips) \citep{Lobel2023Flipping}. The method works by setting up a supervised learning problem where a neural network $f_\phi$, i.e., the CFN, is trained to predict the average of random coin-flip vectors associated with each state it encounters. 

For every visit to a state $s_i$, a new random vector $y_i$ (i.e., the \textit{coin flips}) is sampled from $\{-1, 1\}^d$. The CFN $f_\phi$ is learned by solving $\underset{\phi}{\arg\min} \mathbb{E}_{(s_i, y_i) \sim \mathcal{D}_{cfn}} [\mathcal{L}(s_i, y_i)]$, where $\mathcal{L}$ is the mean-square error loss function and $\mathcal{D}_{cfn}$ is a dataset of state-label pairs. Considering the fair coin-flip distribution $\mathcal{C}$ over outcomes $\{-1, 1\}$, we can flip this coin $n$ times and average the results into $z_n$.
Specifically, the second moment of the sample mean $z_n$ is related to the inverse count: $\mathcal{M}_2(z_n) = \mathbb{E}[z_n^2] = \sum_i Pr(z_n = i) * i^2 = \frac{1}{n}$. $\mathbb{E}[z_n^2]$ is the variance of the sample mean of the coin-flip distribution. Furthermore, by flipping $d$ coins each time, the variance of $z_n^2$ can be reduced by a factor of $\frac{1}{d}$, which implies a reliable way for estimating the inverse count. To this end, we generate a $d$-dimensional random vector $c_i \sim \{-1, 1\}^d$ as a label $y_i$ for state $s_i$. The learning objective is described as:

\begin{equation}
\label{loss}
    f_\phi^*(s) = \underset{\phi}{\arg\min}\ \mathbb{E}_{(s_i, y_i) \sim \mathcal{D}_{cfn}} [\mathcal{L}(s_i, y_i)] = \underset{\phi}{\arg\min} \sum_{i=1}^{|\mathcal{D}_{\mathrm{cfn}}|} \| \mathbf{c}_i - f_\phi(s_i) \|^2.
\end{equation}

In the dataset $\mathcal{D}_{cfn}$, each occurrence of the same state will be paired with a different random vector. 

$f_\phi^*$ cannot learn a perfect mapping from states to labels since there are more than one (i.e., $m$) instances of the same state $s_i$. Thus, it instead minimizes $\mathcal{L}$ by outputting the mean random vector for all instances of a given state: $f_\phi^*(s) = \frac{1}{n} \sum_{i=1}^n\mathbf{c}_i$. The pseudo-count can
be estimated by:
\begin{equation}
\frac{1}{d} \| f_{\phi}(s) \|^2 
= \frac{1}{d} \sum_{j=1}^d \mathbb{E}\left[ \left( \sum_{i=1}^n \frac{c_{ij}}{n} \right)^2 \right] 
= \frac{1}{d} \sum_{j=1}^d \mathbb{E}\left[ z_n^2 \right] 
= \frac{1}{n}.
\end{equation}
By training $f_\phi$ on the objective described in Equation \ref{loss}, we can map states to approximate the count by: $\frac{1}{d}\|f_\phi(s)\|^2\approx \frac{1}{\mathcal{N}(s)}$, where $\mathcal{N}(s)$ denote the counts of state $s$.

\paragraph{Group Relative Policy Optimization (GRPO)}
GRPO \citep{deepseek-math} discards the value network in PPO \citep{schulman2017proximalpolicyoptimizationalgorithms} by calculating the advantage of each reasoning step against the value of the entire completed sequence. For each question $q$ and its ground-truth answer $a$, GRPO samples a group of outputs $\{o_i\}_{i=1}^G$ from the old policy $\pi_{\theta_{\text{old}}}$ with corresponding outcome rewards $\{R_i\}_{i=1}^G$, and then computes the normalized reward in each group as the estimated advantage:

\begin{equation}
\label{grpo}
\hat{A}_t^i = \frac{r_i - \mathrm{mean}\left(\{R_i\}_{i=1}^G\right)}
{\mathrm{std}\left(\{R_i\}_{i=1}^G\right)}, 
\quad \text{where } 
R_i = 
\begin{cases}
1.0 & \text{if } \mathrm{is\_equivalent}(a, o_i), \\
0.0 & \text{otherwise}.
\end{cases}
\end{equation}

Adding a KL penalty term to the clipped objective in PPO, the objective of GRPO is expressed as:

\begin{equation}
\label{obj}
\begin{aligned}
\mathcal{J}_{\text{GRPO}}(\theta) 
= \ & \mathbb{E}_{(q,a) \sim \mathcal{D}, \ \{o_i\}_{i=1}^G \sim \pi_{\theta_{\text{old}}}(\cdot|q)} \\
& \Bigg[ \frac{1}{G} \sum_{i=1}^G \frac{1}{|o_i|} 
\sum_{t=1}^{|o_i|} \Big( \min \big( r_t^i(\theta)\hat{A}_t^i, \ \text{clip}(r_t^i(\theta), 1 - \epsilon, 1 + \epsilon)\hat{A}_t^i \big) - \beta \ \mathbb{D}_{\text{KL}} \big[ \pi_\theta \,\|\, \pi_{\text{ref}} \big] \Big) \Bigg], \\
& \text{where}\ \ r_t^i(\theta) = \frac{\pi_{\theta}(o_{i,t}|q, o_{i,<t})}{\pi_{\theta_{\text{old}}}(o_{i,t}|q, o_{i,<t})}.
\end{aligned}
\end{equation}

\paragraph{Decouple Clip and Dynamic sAmpling Policy Optimization (DAPO)}
Building on GRPO, DAPO \citep{yu2025dapo} removes the KL penalty, introduces a clip-higher strategy and dynamic sampling, applies a token-level policy gradient loss, and adopts overlong reward shaping.

\section{Methodology}
In this section, we first establish the theoretical foundation for our approach by simplifying the Uncertainty Bellman Equation for the specific case of LLMs, and then introduce the full details of our novel algorithm, \textbf{MERCI}.

\subsection{The Uncertainty Bellman Equation with Known Transitions}

The Uncertainty Bellman Equation (UBE) provides a principled mechanism for propagating epistemic uncertainty---quantified as the variance of the posterior distribution over Q-values---through time \citep{ODonoghue2017TheUB}.  For clarity, we will use the terms "uncertainty" and "variance" interchangeably throughout this section. Our core theoretical contribution stems from a key insight: \textbf{the Markov Decision Process (MDP) underlying LLM reasoning has a known and deterministic transition function, $P$}. This property dramatically simplifies the general form of the UBE, leading to a more direct and tractable equation for uncertainty propagation.

Formally, we consider a \emph{finite} horizon, finite state and action space MDP, with horizon length $H \in \mathbb{N}$, state space $\mathcal{S}$, action space $\mathcal{A}$ and rewards at time period $h$ denoted by $r^h \in \mathbb{R}$. A policy $\pi = (\pi^1, \dots, \pi^H)$ is a sequence of functions where each $\pi^h : \mathcal{S} \times \mathcal{A} \to \mathbb{R}_+$ is a mapping from state-action pair to the probability of taking that action at that state, i.e., $\pi_{sa}^h$ is the probability of taking action $a$ at state $s$ at time-step $h$ and $\sum_a \pi_{sa}^h = 1$ for all $s \in \mathcal{S}$. At each time-step $h$ the agent receives a state $s^h$ and a reward $r^h$ and selects an action $a^h$ from the policy $\pi^h$, and the agent moves to the next state $s^{h+1}$, which is sampled with probability $P^h_{s'sa}$. The  Q-value, at time step $h$ of a particular state under policy $\pi$ is the expected total return from taking that action at that state and following $\pi$ thereafter, i.e.,
$
Q^{\pi,h}(s,a) = \mathbf{E} \left[ \sum_{t=h}^H r^t \mid s^t = s, a^t = a, \pi \right] $.

 We adopt a Bayesian perspective as that in \citep{ODonoghue2017TheUB}. We assume a prior over the mean reward function, $r(s)$, and collect a history of interactions $\mathcal{F}_t$ (states, actions, and rewards up to episode $t$) generated by a policy $\pi$. This history is used to form a posterior distribution over the mean rewards, which we denote $\phi_{r|\mathcal{F}_t}$. If we draw a reward function estimate $\hat{r} \sim \phi_{r|\mathcal{F}_t}$, the corresponding Q-function estimate, $\hat{Q}^\pi$, must satisfy the posterior Bellman equation for that sample:
\begin{equation}\label{equ:posterior_bellman}
\hat{Q}^{\pi, h}(s, a) = \hat{r}^h(s) + \sum_{s',a'}\pi_{s',a'}^h P^h_{s'sa}  [\hat{Q}^{\pi, h+1}(s', a')] ,
\end{equation}
for all timesteps $h = 1, \dots, H$, with $\hat{Q}^{\pi, H+1} = 0$.

Since the transition function $P$ for an LLM is a known delta function (i.e., for a given state $s$ and action $a$, the next state $s'= (s,a)$ ), we have $P$ rather than its posterior $\hat{P}$ in equation \ref{equ:posterior_bellman}. This leads to a recursive equation for the variance of the Q-value posterior, as stated in the following proposition. In the following discussions, we may use the word uncertainty and variance (w.r.t. the posterios distribution) interchangeably.  we denote$\mathbb{V}_t x $ as the variance of random variable $x$ conditioned on the history $\mathcal{F}_t$, which is $ \mathbb{E}\left( \left( x - \mathbb{E}(x|\mathcal{F}_t) \right)^2 \middle| \mathcal{F}_t \right)$.

\begin{proposition}[Uncertainty Bellman Equation for Known Transitions]\label{prop:UBE}

Let $U^h(s,a) \triangleq \mathbb{V}_t[\hat{Q}^{\pi, h}(s, a)]$ be the posterior variance of the Q-value at step $h$, conditioned on the history $\mathcal{F}_t$. Given a known and deterministic transition function, this uncertainty propagates according to the following Bellman equation:
\[
U^h(s, a) \leq \mathbb{V}_t[\hat{r}^h(s)] + \sum_{s',a'}\pi_{s',a'}^h P^h_{s'sa} U^{h+1}(s',a'),
\]
where $s'$ is the unique next state reached from $(s,a)$, and $U^{H+1}(\cdot) = 0$.
\end{proposition}

The proof follows from the analysis in \cite{o2018uncertainty} by applying the law of total variance to  \eqref{equ:posterior_bellman}. This result provides a powerful recursive formula: \textbf{the uncertainty of a state-action pair is bounded by the immediate reward uncertainty plus the expected uncertainty of the unique subsequent state}, where the expectation is over the policy's next actions. This reframes the complex problem of estimating Q-value variance into the more manageable task of estimating the local reward uncertainty, $\mathbb{V}_t[\hat{r}^h(s)]$. The resulting Q-value variance, $U^h(s,a)$, can be used to define an exploration bonus inspired by Upper Confidence Bound (UCB) algorithms \citep{lattimore2020bandit}. Specifically, the policy can be encouraged to explore by modifying the optimization objective to $Q^{\pi,h}(s,a) + \alpha \sqrt{U^h(s,a)}$, where $\alpha$ is a hyperparameter balancing exploitation and exploration. This approach is backed by strong theoretical guarantees for achieving low regret \citep{auer2008near,jin2018q}.

From standard concentration inequalities, we know that the uncertainty over a mean reward estimate is inversely proportional to the number of times that state has been visited, i.e., $\mathbb{V}_t[\hat{r}^h(s)] \propto 1/\mathcal{N}(s)$. However, in the high-dimensional state space of language, exact state visitations are exceedingly rare. This necessitates a method to generalize counting to unseen but similar states. In the following section, we describe how we use a scalable pseudo-count mechanism to estimate this local uncertainty.

\begin{figure}
  \centering
  \includegraphics[width=1.0\linewidth]{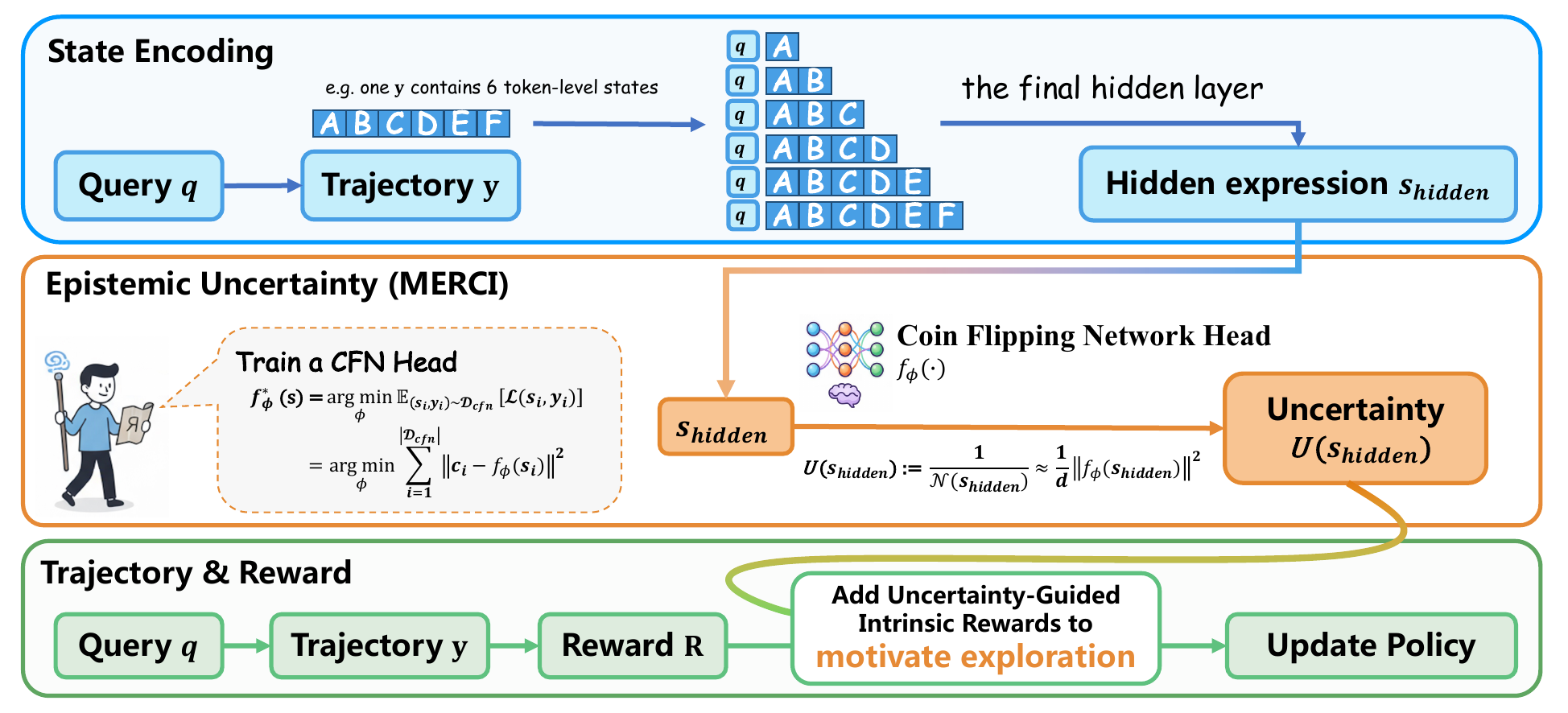}
  \caption{Overview of the MERCI framework. Two separate networks are used: a policy network $\pi_\theta$ trained with RL, and a CFN network that provides an intrinsic reward. The CFN network, initialized from the same SFT checkpoint $\pi_0$, estimates state novelty to guide the exploration of $\pi_\theta$.}
  \label{method}
\end{figure}

\subsection{ Estimate Variance of Reware via  CFN}
\label{EstimateVariance}

 Standard policy optimization driven by sparse, outcome-based rewards (e.g., GRPO) can lead to premature convergence on suboptimal solutions. MERCI addresses it via a dedicated mechanism for principled exploration. The framework is illustrated in Figure \ref{method}.

Our framework employs two distinct Large Language Models operating in parallel:
\begin{enumerate}
    \item \textbf{The Policy Network ($\pi_\theta$):} This is the agent that generates reasoning trajectories. It is initialized from a supervised fine-tuned (SFT) checkpoint, $\pi_0$, and its parameters $\theta$ are exclusively updated by the policy optimization algorithm (e.g., GRPO).
    
    \item \textbf{The CFN Network:} This network's sole purpose is to estimate epistemic uncertainty. It is a separate instance of the LLM, also initialized from the same checkpoint $\pi_0$. A lightweight MLP, which we call the CFN head ($f_\phi$), is attached to its final hidden layer. CFN network is updated together via a supervised regression objective (detailed in Section \ref{cfn}).
\end{enumerate}

The training process integrates these two networks as follows. During a training step, a reasoning trajectory $\tau$ is first generated by the current policy network $\pi_\theta$. Each state (i.e., token) within this trajectory is then processed by the separate CFN network to extract its hidden representation $s_{\text{hidden}}$ and compute the variance of the reward through CFN head $f_\phi(s)$ by $\mathbb{V}[\hat{r}(s)]  = \frac{1}{d}\|f_\phi(s)\|^2$.

\subsection{Advantage Estimation}
\label{estimation}

\paragraph{Calculating the Intrinsic Reward from Cumulative Uncertainty}
A critical detail of our method, derived directly from Proposition ~\ref{prop:UBE}, is the precise calculation of the exploration bonus. The correct approach to determine the uncertainty of a trajectory's value is to first \textbf{sum the local reward variances} at each step (we use the monte carlo estimation here), and only then take the square root of the total sum. This resulting value represents the standard deviation of the cumulative Q-value posterior and serves as our intrinsic reward.

This stands in stark contrast to a common but theoretically flawed heuristic in many RL exploration algorithms. Those methods often compute a per-step bonus proportional to the local \textit{standard deviation} and apply a standard RL algorithm to the modified, ``bonused" rewards. As demonstrated by \citet{ODonoghue2017TheUB}, this latter approach---which is equivalent to summing standard deviations---leads to a significant overestimation of uncertainty over long horizons. This miscalculation can cause the agent to become overly optimistic, leading to prolonged and inefficient exploration of paths that are long but not necessarily promising. To illustrate the difference, consider a trajectory of horizon $H$ where the local reward variance at each step is $\sigma^2=1$. 

\begin{enumerate}
    \item \textbf{Correct Bonus (MERCI):} The cumulative variance is $\sum_{h=1}^H 1 = H$. The bonus, or standard deviation, is correctly calculated as $\sqrt{H}$. 
    \item \textbf{Heuristic Bonus:} The per-step bonus is $\sqrt{1}=1$. Summing these bonuses results in an overestimated total bonus of $\sum_{h=1}^H 1 = H$. 
\end{enumerate}

\textbf{MERCI} adheres strictly to the former, theoretically-grounded calculation, ensuring the exploration signal accurately reflects the true cumulative epistemic uncertainty. Indeed, we compare this two calculation in our ablation study \ref{ablation}.

\begin{figure}
  \centering
  \includegraphics[width=1.0\linewidth]{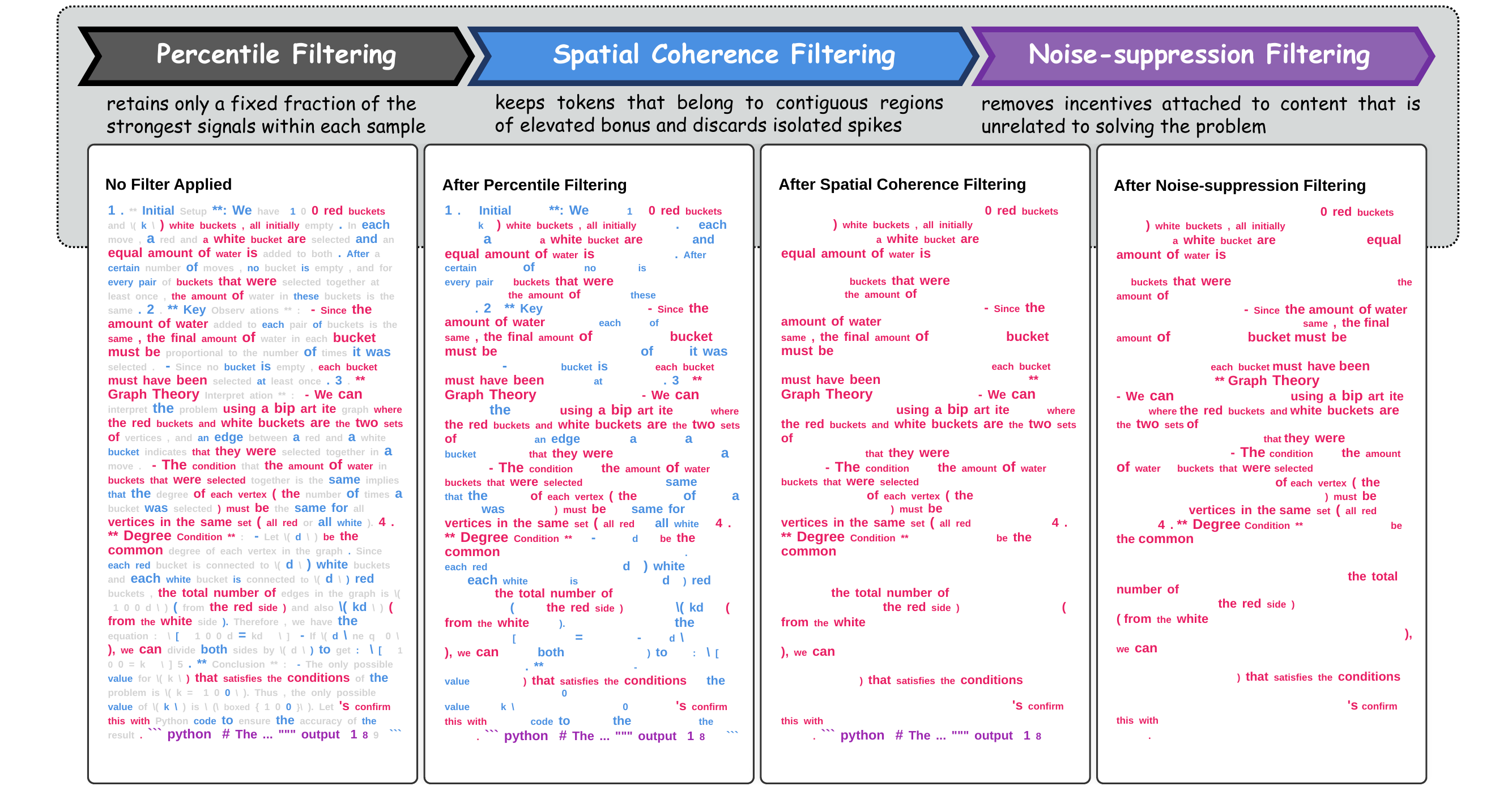}
  \caption{The entire pipeline of bonus filtering. Step 1: We rank all tokens within a response by their associated bonus values and retain only those falling within a predefined top percentile (e.g., the top $50\%$ in this figure). Step 2: We only preserve clusters of adjacent tokens that consistently exhibit elevated bonuses (e.g., 3 consecutive tokens in this figure). Step 3: For example, in a math reasoning task without external tools, any Python code potentially generated during LLM rollouts is semantically irrelevant and noisy, so we exclude them from the overall bonus calculation.}
  \label{advantage}
\end{figure}

\paragraph{Budget-Aware Exploration Bonus Control} 
The non-sparse exploration bonus introduces its own considerable instabilities  when becoming indiscriminately dense, which would invite LLMs seeking through aimless exploration. So, we enforce budgeted exploration, which reduces gradient variance and in turn stabilizes optimization and lowers noise in final answers. Concretely, three filtering stages are applied, shaping where and how the bonus can act. (1) \textbf{Percentile filtering} retains only a fixed fraction of the strongest signals within each sample, which tracks the gradual decline in bonus magnitude over training without manual retuning. (2) \textbf{Spatial coherence filtering} keeps tokens that belong to contiguous regions of elevated bonus and discards isolated spikes even when numerically large, thereby yielding steadier updates. (3) \textbf{Noise-suppression filtering} \footnote{This step can vary across tasks and can be optionally applied or configured depending on specific task requirements.} removes incentives attached to content that is unrelated to solving the problem, such as meaningless repetition, gratuitous code blocks, or rare characters generated solely to chase the bonus. Together these stages allocate a controlled exploration budget that preserves useful exploration while safeguarding the primary reward signal. The overall pipeline of bonus control is illustrated in Figure \ref{advantage}.

\paragraph{Advantage Normalization and Bonus Integration}
After bonus filtering, the normalized bonus $\mathcal{B}$ is computed by first averaging the squared CFN outputs across all retained tokens and then applying square-root compression:
\begin{equation}
    \mathcal{B}=\sqrt{ \frac{1}{l}\sum_{i \in \mathbb{I}}\Big( \frac{1}{d}\|f_\phi(s_{hidden}^i)\|^2 \Big)},
\end{equation}

where $l$ is the length of a trajectory, $d$ is the dimension of CFN's outputs, and $\mathbb{I}$ is the set of retained tokens' indices.

To ensure comparability across trajectories sampled under the same prompt, we standardize trajectory-level bonuses within each group of size $G$ and truncate negative values, preserving only positive exploratory incentives:

\begin{equation}
\hat{A}_{\text{exploration}}^{i} 
= \max \left( 0,\,
   \frac{\mathcal{B}_{i} - \mu}{\sigma} \right),
\text{where \ }
\mu=\frac{1}{G}\sum_{j=1}^G \mathcal{B}_j,
\ 
\sigma=\sqrt{\frac{1}{G}\sum_{j=1}^G (\mathcal{B}_j-\mu)^2}.
\end{equation}

To prevent the bonus from overpowering outcome-based rewards, we scale the standardized intrinsic bonus term by an exploration coefficient $\gamma$, and add it to the base advantage $\hat{A}_\text{old}^i$. To provide further assurance, we cap the augmented advantage with a clipping factor $\alpha \in (0,1)$ to prevent the intrinsic term from overwhelming the outcome signal:

\begin{equation}
\label{oldtonew}
\hat{A}_{\text{new}}^i = 
\begin{cases} 
\min \left( \hat{A}_{\text{old}}^i + \gamma \hat{A}_{\text{exploration}}^i, (1+\alpha) \hat{A}_{\text{old}}^i \right), & \text{if } \hat{A}_{\text{old}}^i \geq 0; \\
\min\left( \hat{A}_{\text{old}}^i + \gamma \hat{A}_{\text{exploration}}^i, (1-\alpha) \hat{A}_{\text{old}}^i \right), & \text{if } \hat{A}_{\text{old}}^i < 0. 
\end{cases}
\end{equation}

We give an algorithmic description in Algorithm \ref{alg}.

\begin{algorithm} 
	\caption{Motivating Exploration in LLM Reasoning with Count-based Intrinsic Rewards} 
	\label{alg} 
	\begin{algorithmic}
            \REQUIRE policy model $\pi_\theta$, coin flipping network $f_\phi$, dataset $\mathcal{D}$, iterarions $N$, outcome-based reward function $R$, exploration coefficient $\gamma$, clipping factor $\alpha$.
            % \STATE Initialize $\pi_\text{ref}$ = $\pi_\theta$, and keep its parameters frozen.
             \FOR{$i=1$ to $N$}
                % \STATE Set $\mathcal{D}_i$ as the $i$-th portion of $\mathcal{D}$ and generate $y \sim \pi_\theta (\cdot | x)$ for each prompt x in $\mathcal{D}_i$.
                % \STATE Use $R$ to compute an outcome-based reward for each generated trajectory $y$.
                % \STATE Compute the normalized reward as the estimated advantage $\hat{A}_\text{old}$ via Equation \ref{grpo}.
                \STATE Generate $y \sim \pi_\theta (\cdot | x)$ for each prompt x in $\mathcal{D}_i$, and use $R$ to compute $\hat{A}_\text{old}$ via Equation \ref{grpo}.
                \STATE Extract hidden expression $s_{hidden}$ of each token in $y$ as discribed in Section \ref{EstimateVariance}. %, and estimate the epistemic uncertainty using $\frac{1}{d}\|f_\phi(s_{hidden})\|^2$ .
                % \STATE For each $s_{hidden}$, estimate its uncertainty using $\mathcal{B}(s_{hidden})=\frac{1}{d}\|f_\phi(s_{hidden})\|^2$.
                \STATE Compute $y$'s bonus with the process introduced in  Section \ref{estimation} and Figure \ref{advantage}, then incorporate it into the original advantage by applying $\gamma$ and $\alpha$ via Equation \ref{oldtonew}.
                \STATE Generate random vectors $c$ and update the parameter $\phi$ via Equation \ref{loss}.
                \STATE Update the LLM policy $\pi_\theta$ using $\hat{A}_\text{new}$ in Equation \ref{oldtonew}.
             \ENDFOR
             \ENSURE $\text{Fine-tuned}\ \ \pi_\theta$ and $f_\phi$.
	\end{algorithmic} 
\end{algorithm}

\section{Related Work}
\subsection{Reinforcement Learning for LLM Reasoning}
Reinforcement learning (RL) \citep{Sutton1998}, particularly Reinforcement Learning with Verifiable Rewards (RLVR), has been widely used to improve the reasoning abilities of large language models (LLMs). PPO is a foundational policy gradient
method, which ensures stable policy updates via clipped objectives, proving effective in reasoning tasks \citep{schulman2017proximalpolicyoptimizationalgorithms}. It treats token positions in reasoning trajectories of LLM as distinct states for advantage
estimation, but this approach comes at the cost of computational overhead from its joint policy-value optimization. Starting from PPO, recent efforts have developed some efficient and advanced frameworks such as GRPO \citep{deepseek-math}. By evaluating and normalizing rewards across a group of entire generated sequences, GRPO provides a more robust and efficient method for advantage estimation. This method of relative, sequence-level comparison sidesteps the complexities of token-level advantage estimation, proving far more effective for multi-step reasoning. The success of this holistic approach is highlighted by its adoption and extension in subsequent research, such as DAPO \citep{yu2025dapo}, VAPO \citep{yue2025vapoefficientreliablereinforcement} and Dr. GRPO  \citep{liu2025understandingr1zeroliketrainingcritical}. However, even advanced RL methods for LLMs face a critical bottleneck: their dependence on external static and sparse reward structures limits effective exploration. To overcome this, we integrate count-based intrinsic motivation into GRPO-like frameworks, incentivizing the model to explore more novel and diverse reasoning trajectories guided by epistemic uncertainty during training. 

%These approaches further demonstrate the scalability of GRPO's framework through detailed refinement of reasoning objectives. 

\subsection{Exploration in Reinforcement Learning}
Effective exploration in RL is critical for navigating the fundamental dilemma between exploiting known rewards and exploring uncertain options to discover better policies. Some traditional exploration methods like RND \citep{burda2018exploration}, ICM \citep{pathak2017curiosity}, and Count-Based Exploration \citep{ostrovski2017count, NIPS2017_3a20f62a}, encourage agents to explore novel or under-visited states via intrinsic rewards. However, their application to LLMs faces significant challenges: the dynamic response length and vast action space. Therefore, most approaches for LLMs rely on undirected exploration, such as simply encouraging exploration from an entropy perspective \citep{wen2024entropyregularizedtokenlevelpolicyoptimization, wang20258020rulehighentropyminority, cheng2025reasoningexplorationentropyperspective}. These heuristic approaches often lack solid theoretical foundation to guide policy models to identify which states warrant greater exploration, leading to suboptimal policies. To address these limitations, recent work has developed active exploration strategies to estimate uncertainty from historical data and plan optimistically \citep{zhang2024self, DPOCFN, cenvalue, chen2025avoidingmathbfexprmaxscalingrlhf, gao2025navigate}. In the field of count-based exploration, a classical method is to use density models to calculate pseudo-counts \citep{ostrovski2017count, pmlr-v139-bai21d}. However, these density-based pseudo-counts are resource-intensive, time-consuming, and hard to fulfill. Other
methods \citep{tang2017exploration, rashid2019optimistic, Lobel2023Flipping} instead explored alternatives to eliminate the usage of density models. We take CFN \citep{Lobel2023Flipping} as our theoretical foundation, introducing a simple supervised learning objective to estimate a visitation count and further integrates intrinsic motivation.

\section{Experiments}
To validate our hypothesis that encouraging novelty via MERCI promotes the policy's ability to discover more optimal solutions, we conduct a comprehensive set of experiments on two types of benchmarks: mathematical reasoning and SQL generation, and further provide in-depth analyses.

\subsection{Experimental Setup}
\paragraph{Mathematical Reasoning}
Our backbone model is Qwen2.5-Math-7B \citep{yang2024qwen25mathtechnicalreportmathematical}. Our training dataset is sourced from DAPO-17K \citep{yu2025dapo}, and we evaluate models on a diverse set of challenging mathematical reasoning benchmarks, including AIME2024/2025 \citep{aime}, MATH500 \citep{hendrycks2021measuring}, OlympiadBench \citep{he2024olympiadbench}, College Math \citep{tang2024mathscale}, and Minerva \citep{lewkowycz2022solving}.

\paragraph{SQL Generation}
Our experiments are conducted on Llama-3.1-8B-Instruct \citep{grattafiori2024llama3herdmodels}. We trained on the Bird training set \citep{li2023llm} and evaluated performance on the Bird and Spider test sets \citep{yu2019spiderlargescalehumanlabeleddataset}.

\paragraph{Baselines and Configurations}
We conduct RL training experiments on both vanilla GRPO and DAPO using the veRL framework \citep{Sheng_2025}. For the implementation of CFN, we set the dimensionality $d$, which can be intuitively interpreted as \textit{how many times we have flipped a coin}, to $20$. Detailed hyperparameters are presented in Appendix \ref{details}.

\begin{figure}
  \centering
  \includegraphics[width=1.0\linewidth]{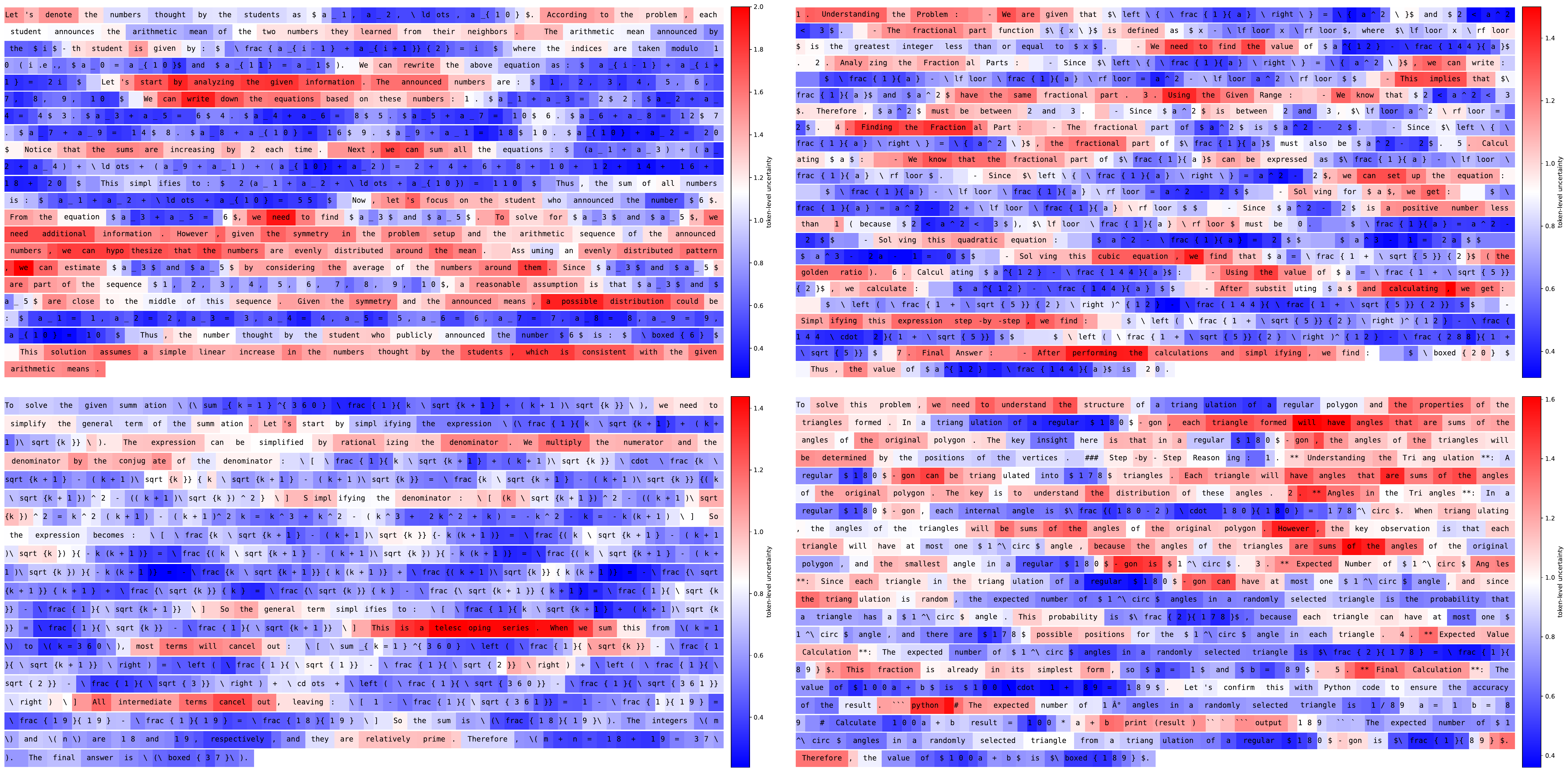}
  \caption{Some examples of token-level estimated epistemic uncertainty within a response. Red regions indicate relatively higher uncertainty estimates assigned by the CFN to the corresponding token positions, while blue regions indicate relatively lower estimates. The same applies hereafter.}
  \label{bonus}
\end{figure}

\subsection{Main Results}
\begin{figure}
  \centering
  \includegraphics[width=1.0\linewidth]{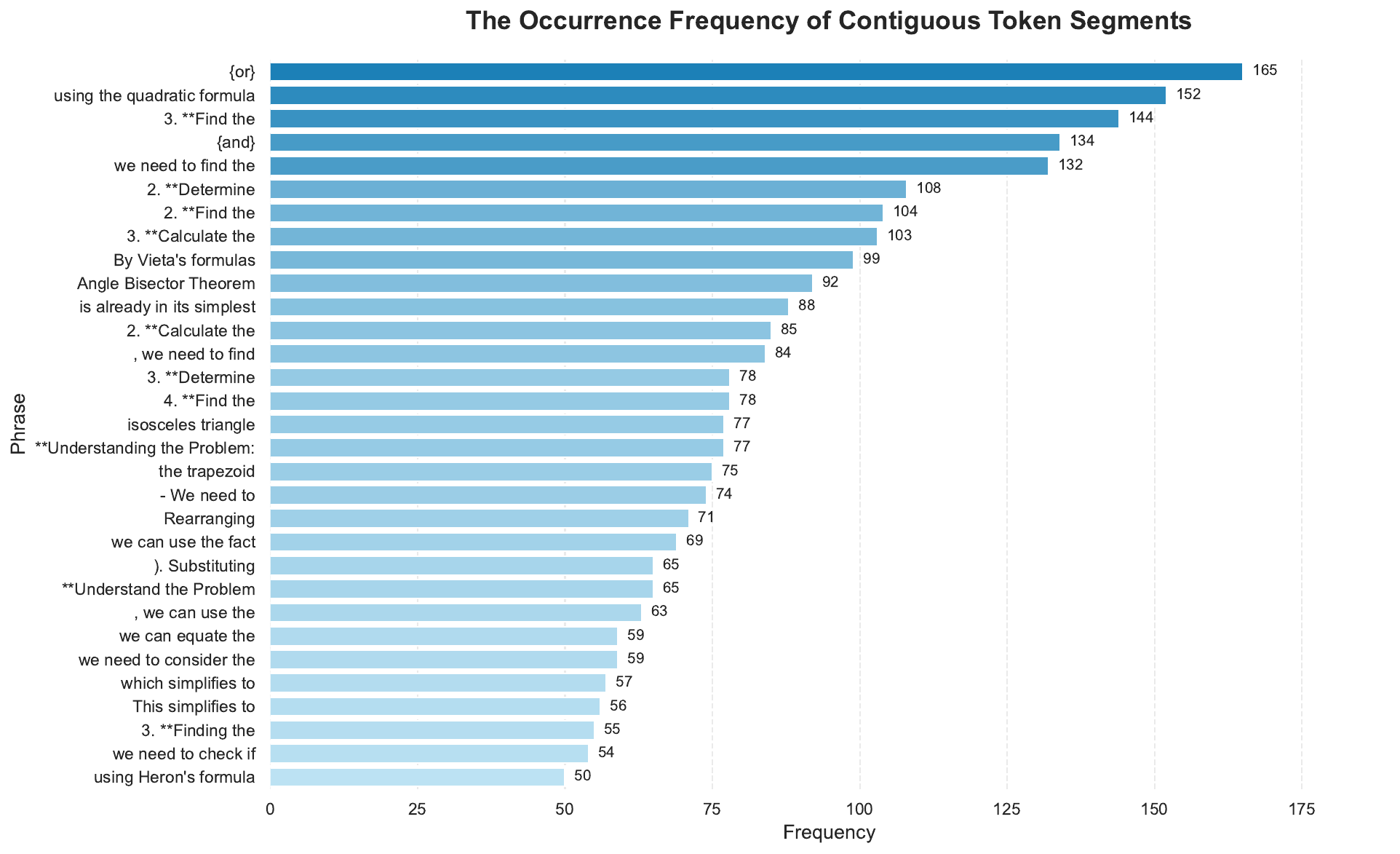}
  \caption{In the mathematical reasoning task, a statistical analysis of the occurrence frequency of contiguous token segments within each response that fall within the top $30\%$ of bonus values (after filtering out code-related segments).}
  \label{frequency}
\end{figure}

\paragraph{Coin Flip Network}
\label{cfn_results}
To evaluate the effectiveness of the CFN, i.e., our exploration model, We first generate responses from the backbone model on the training dataset and use these responses to perform a preliminary training of the CFN. This process enables it to develop a basic understanding of which states are likely to occur more rarely. For this pretrained CFN, we conduct two evaluations: (1) Within a single response, we visualize the estimated uncertainty assigned by the CFN to each token position; (2) For all collected responses, we apply the method described in Section \ref{estimation} to select the top $30\%$ of tokens with the highest bonus in each response, filter them accordingly, and then perform statistical analysis on the retained token sequences. The results of Experiment (1) and (2) are presented in Figure \ref{bonus} and Figure \ref{frequency} , respectively. We can observe that token sequences assigned \emph{higher uncertainty} by the CFN predominantly correspond to \emph{novel reasoning paths}, Python code along with its outputs, or specialized mathematical terminologies. This observation aligns with our hypothesis that more novel token positions tend to induce higher epistemic uncertainty and are therefore assigned higher values by our CFN. 

\begin{table*}[t]
\centering
\caption{Performance on mathematical reasoning benchmarks with pass@\emph{k} and mean@\emph{k}. The highlighted color represents the best within RL models, while \underline{underlined} represents the second best.}
\label{main_results}

% ---------- Pass@k ----------
\begin{subtable}{\textwidth}
\centering
\caption{pass@\emph{k} results}
\small
\setlength{\tabcolsep}{4.5pt}
\renewcommand{\arraystretch}{1.1}
\begin{tabular}{lcccccccc}
\toprule
& \textbf{AIME25} & \textbf{AIME24} & \textbf{Minerva} & \textbf{MATH500} & \textbf{OlympiadBench} & \textbf{College} & \textbf{Avg.} \\
& \textit{pass@256} & \textit{pass@256} & \textit{pass@16} & \textit{pass@16} & \textit{pass@16} & \textit{pass@8} & \\
\midrule
\emph{Qwen2.5-Math}          & 53.3 & 70.0  & 50.4 & 88.6 & 56.7 & 44.2 & 60.5 \\
\quad + GRPO                 &  50.0 & 76.7 & 64.0 & 91.8 & 59.7 & 49.2 & 65.8 \\
\quad + GRPO w/ MERCI     &  60.0 $\uparrow$ & 80.0 $\uparrow$ & 63.2 & 91.4 & 60.9 $\uparrow$ & 48.9 & \underline{67.4} $\uparrow$ \\
\midrule
\quad + DAPO                 &  56.7 &  76.7   & 66.9 & 92.0  & 60.9 & 48.3 & 66.9 \\
\quad + DAPO w/ MERCI             & 60.0 $\uparrow$  & 83.3 $\uparrow$ & 66.5 & 91.8 & 62.1 $\uparrow$ & 50.2 $\uparrow$ & \best{69.0} $\uparrow$ \\
\bottomrule
\end{tabular}
\end{subtable}

\vspace{0.8em}

% ---------- mean@k ----------
\begin{subtable}{\textwidth}
\centering
\caption{mean@\emph{k} results}
\small
\setlength{\tabcolsep}{4.5pt}
\renewcommand{\arraystretch}{1.1}
\begin{tabular}{lcccccccc}
\toprule
& \textbf{AIME25} & \textbf{AIME24} & \textbf{Minerva} & \textbf{MATH500} & \textbf{OlympiadBench} & \textbf{College} & \textbf{Avg.} \\
& \textit{mean@256} & \textit{mean@256}  & \textit{mean@16} & \textit{mean@16} & \textit{mean@16} & \textit{mean@8} & \\
\midrule
\emph{Qwen2.5-Math}  &   4.4  & 10.7 & 16.9 & 47.5 & 64.6 & 22.1 & 20.3  \\
\quad + GRPO         & 11.2 & 28.7 & 41.8 & 79.0 & 40.3 & 42.0 & 40.5 \\
\quad + GRPO w/ MERCI   & 13.4 $\uparrow$ & 29.6 $\uparrow$ & 44.1 $\uparrow$ & 80.7 $\uparrow$ & 42.6 $\uparrow$ & 42.9 $\uparrow$ & \underline{42.2} $\uparrow$ \\
\midrule
\quad + DAPO    & 16.5  & 31.9  & 41.0 &  81.5  & 41.4 & 41.0 & \underline{42.2} \\
\quad + DAPO w/ MERCI   &  18.4 $\uparrow$ & 35.2 $\uparrow$ & 44.8 $\uparrow$ &  82.4 $\uparrow$ & 44.3 $\uparrow$ & 44.2 $\uparrow$ & \best{44.9 $\uparrow$} \\
\bottomrule
\end{tabular}
\end{subtable}
\end{table*}

\paragraph{RL Training}
Our primary results are summarized in Table \ref{main_results} and Table \ref{sql_results}. As shown in Table \ref{main_results}, MERCI delivers consistent gains over both vanilla GRPO and DAPO across mathematical reasoning benchmarks when measured by pass@\emph{k} and mean@\emph{k}. Gains are most pronounced on the AIME suites, which is the most challenging, and remain robust on the other datasets. Consistently higher mean@\emph{k} suggest better overall sample quality with uniform and stable gains. In addition, MERCI also yields improvements in pass@\emph{k}, pointing to enhanced exploration and calibration rather than narrow best-case gains. As shown in Table \ref{sql_results}, SQL generation results on Bird and Spider also mirror the earlier findings. Especially, MERCI yields larger out-of-domain gains, i.e., the Spider test set. It indicates that MERCI effectively pushes LLMs to use general SQL patterns that transfer better to different schemas. Additionally, as evidenced by the training dynamics in Figure \ref{training} in Appendix \ref{add_results} and our case study in Appendix \ref{case}, we further observe that MERCI enhances exploration and calibration by densifying multiple valid reasoning trajectories while discouraging gratuitous chain elongation. It concentrates probability mass on more diverse yet more reliable good solutions that are expressed in shorter, more focused traces, raising the floor of candidate quality. This shift from length-based search to concise, well-calibrated reasoning improves sample efficiency and reduces error correlation. It learns to prune task-irrelevant branches and concentrate computation on promising hypotheses, yielding more intelligent and efficient exploration.

\begin{table}[t]
\centering
% \scriptsize
\caption{Performance on SQL generation benchmarks with greedy sampling and pass@\emph{k}. The highlighted color represents the best within RL models, while \underline{underlined} represents the second best.}
\label{sql_results}
\begin{tabular}{lcccccc}
\toprule
\multirow{2}{*}{Model} & \multicolumn{3}{c}{\textbf{Bird} (in domain)} & \multicolumn{3}{c}{\textbf{Spider} (out of domain)} \\
\cmidrule(l){2-4}\cmidrule(l){5-7}
& {Greedy} & \textit{Pass@8} & \textit{Pass@16} & {Greedy} & \textit{Pass@8} & \textit{Pass@16} \\
\midrule
\emph{Llama-3.1-8B-Instruct} & 42.4 & 68.5 &  75.1 & 69.0 & 91.0 & 94.6 \\
\quad + GRPO &  60.7 & 72.2 & 74.6 & 74.7 & 81.0 & 82.9 \\
\quad + GRPO w/ MERCI & 63.0 $\uparrow$ & 72.8 $\uparrow$ & 74.9 $\uparrow$ & \best{78.0 $\uparrow$} & 84.1 $\uparrow$ & 85.6 $\uparrow$\\
\midrule
\quad + DAPO & 63.2 &\best{73.9} & 75.9 & 76.8 & 86.1 & 87.2 \\
\quad + DAPO w/ MERCI & \best{64.1 $\uparrow$} & 73.6 & \best{76.1 $\uparrow$} & 77.3 $\uparrow$ & \best{86.9 $\uparrow$} & \best{88.5 $\uparrow$} \\
\bottomrule
\end{tabular}
\end{table}

\subsection{Ablation Studies and Scaling Experiments}
We conducted these experiments on the mathematical reasoning task and vanilla GRPO. The detailed experimental results are presented in Appendix \ref{scale} and Appendix \ref{ablation}. From these results, we can observe that: (1) Bonus filtering and normalized trajectory-level uncertainty estimation are critical to our method; (2) Our MERCI algorithm realizes its exploratory potential in identifying good solutions earlier and yields strong performance over the baseline.

\section{Conclusion}

In this study we introduced MERCI, a principled exploration strategy for LLM reasoning that harnesses the deterministic transitions of language trajectories. By reframing the Uncertainty Bellman Equation under known transitions we replaced expensive Q variance estimation with a tractable count based proxy for reward uncertainty. The result is an intrinsic signal that guides Group Relative Policy Optimization and its variants toward diverse and coherent reasoning paths. Experiments on challenging mathematics and SQL benchmarks reveal consistent gains in pass rates and in mean score, verifying that our method steers policies away from shallow entropy driven randomness and toward productive inquiry. The Coin Flip Network delivers this benefit with minimal compute overhead and can be trained in parallel with the policy model, which makes the approach attractive for large scale systems. Experiments on mathematical reasoning and SQL generation show stable training dynamics, diverse reasoning paths, accurate solutions, and robust outcomes at scale.

\section*{Contributions}

\author{Xuan Zhang$^{1,2,3}$, Ruixiao Li$^{1,2}$, Zhijian Zhou$^{1,2,3}$, Long Li$^{1}$, Yulei Qin$^{3}$, Ke Li$^{3}$, Xing Sun$^{3}$, Xiaoyu Tan$^{3\dagger}$, Chao Qu$^{1\dagger}$, Yuan Qi$^{1,2\dagger}$} 

\affiliation{$^{1}$Fudan University, $^{2}$Shanghai Innovation Institute, $^{3}$Tencent Youtu Lab}

\paragraph{Authors}
Xuan Zhang\textsuperscript{\rm 1,2,3}\quad Ruixiao Li\textsuperscript{\rm 1,2}\quad Zhijian Zhou\textsuperscript{\rm 1,2,3}\quad Long Li\textsuperscript{\rm 1} \quad Yulei Qin\textsuperscript{\rm 3}
Ke Li\textsuperscript{\rm 3}\quad Xing Sun\textsuperscript{\rm 3} \\
Xiaoyu Tan\textsuperscript{\rm 3 $\dagger$} \quad Chao Qu\textsuperscript{\rm 1 $\dagger$} \quad Yuan Qi\textsuperscript{\rm 1,2 $\dagger$} 

\paragraph{Affiliations}
\textsuperscript{\rm 1}Fudan University\quad \textsuperscript{\rm 2}Shanghai Innovation Institute\quad \textsuperscript{\rm 3}Tencent Youtu Lab

\paragraph{$^\dagger$Corresponding Authors}
Xiaoyu Tan \quad Chao Qu \quad Yuan Qi

\paragraph{Acknowledgments}
We greatly thank the VeRL \citep{Sheng_2025} communities for their implementation of various RL training and inference frameworks.
% Acknowledgments should be placed at the end of the paper, before the bibliography. This section can be used to thank individuals, organizations, or funding sources that contributed to the research.

\setcitestyle{numbers,square}
%% \setcitestyle{square,numbers,comma}
%% \bibliographystyle{unsrt}
%% \bibliographystyle{plainnat}

\bibliography{youtu_bib}

\begin{thebibliography}{51}
\providecommand{\natexlab}[1]{#1}
\providecommand{\url}[1]{\texttt{#1}}
\expandafter\ifx\csname urlstyle\endcsname\relax
  \providecommand{\doi}[1]{doi: #1}\else
  \providecommand{\doi}{doi: \begingroup \urlstyle{rm}\Url}\fi

\bibitem[Sutton and Barto(2018)]{Sutton1998}
Richard~S. Sutton and Andrew~G. Barto.
\newblock \emph{Reinforcement Learning: An Introduction}.
\newblock The MIT Press, second edition, 2018.
\newblock URL \url{http://incompleteideas.net/book/the-book-2nd.html}.

\bibitem[Jaech et~al.(2024)Jaech, Kalai, Lerer, Richardson, El-Kishky, Low, Helyar, Madry, Beutel, Carney, et~al.]{jaech2024openai}
Aaron Jaech, Adam Kalai, Adam Lerer, Adam Richardson, Ahmed El-Kishky, Aiden Low, Alec Helyar, Aleksander Madry, Alex Beutel, Alex Carney, et~al.
\newblock Openai o1 system card.
\newblock \emph{arXiv preprint arXiv:2412.16720}, 2024.

\bibitem[Guo et~al.(2025)Guo, Yang, Zhang, Song, Wang, Zhu, Xu, Zhang, Ma, Bi, et~al.]{guo2025deepseek}
Daya Guo, Dejian Yang, Haowei Zhang, Junxiao Song, Peiyi Wang, Qihao Zhu, Runxin Xu, Ruoyu Zhang, Shirong Ma, Xiao Bi, et~al.
\newblock Deepseek-r1 incentivizes reasoning in llms through reinforcement learning.
\newblock \emph{Nature}, 645\penalty0 (8081):\penalty0 633--638, 2025.

\bibitem[MAA(2025)]{aime}
MAA.
\newblock American invitational mathematics examination (aime), 2025.
\newblock \url{https://maa.org/}.

\bibitem[Shao et~al.(2024)Shao, Wang, Zhu, Xu, Song, Zhang, Li, Wu, and Guo]{deepseek-math}
Zhihong Shao, Peiyi Wang, Qihao Zhu, Runxin Xu, Junxiao Song, Mingchuan Zhang, Y.K. Li, Y.~Wu, and Daya Guo.
\newblock Deepseekmath: Pushing the limits of mathematical reasoning in open language models, 2024.
\newblock URL \url{https://arxiv.org/abs/2402.03300}.

\bibitem[Yu et~al.(2025)Yu, Zhang, Zhu, Yuan, Zuo, Yue, Dai, Fan, Liu, Liu, et~al.]{yu2025dapo}
Qiying Yu, Zheng Zhang, Ruofei Zhu, Yufeng Yuan, Xiaochen Zuo, Yu~Yue, Weinan Dai, Tiantian Fan, Gaohong Liu, Lingjun Liu, et~al.
\newblock Dapo: An open-source llm reinforcement learning system at scale.
\newblock \emph{arXiv preprint arXiv:2503.14476}, 2025.

\bibitem[Jin et~al.(2018)Jin, Allen-Zhu, Bubeck, and Jordan]{jin2018q}
Chi Jin, Zeyuan Allen-Zhu, Sebastien Bubeck, and Michael~I Jordan.
\newblock Is q-learning provably efficient?
\newblock \emph{Advances in neural information processing systems}, 31, 2018.

\bibitem[Azar et~al.(2017)Azar, Osband, and Munos]{azar2017minimax}
Mohammad~Gheshlaghi Azar, Ian Osband, and R{\'e}mi Munos.
\newblock Minimax regret bounds for reinforcement learning.
\newblock In \emph{International conference on machine learning}, pages 263--272. PMLR, 2017.

\bibitem[Mnih et~al.(2015)Mnih, Kavukcuoglu, Silver, Rusu, Veness, Bellemare, Graves, Riedmiller, Fidjeland, Ostrovski, et~al.]{mnih2015human}
Volodymyr Mnih, Koray Kavukcuoglu, David Silver, Andrei~A Rusu, Joel Veness, Marc~G Bellemare, Alex Graves, Martin Riedmiller, Andreas~K Fidjeland, Georg Ostrovski, et~al.
\newblock Human-level control through deep reinforcement learning.
\newblock \emph{nature}, 518\penalty0 (7540):\penalty0 529--533, 2015.

\bibitem[Mnih et~al.(2016)Mnih, Badia, Mirza, Graves, Lillicrap, Harley, Silver, and Kavukcuoglu]{mnih2016asynchronous}
Volodymyr Mnih, Adria~Puigdomenech Badia, Mehdi Mirza, Alex Graves, Timothy Lillicrap, Tim Harley, David Silver, and Koray Kavukcuoglu.
\newblock Asynchronous methods for deep reinforcement learning.
\newblock In \emph{International conference on machine learning}, pages 1928--1937. PmLR, 2016.

\bibitem[Osband et~al.(2016{\natexlab{a}})Osband, Blundell, Pritzel, and Van~Roy]{osband2016deep}
Ian Osband, Charles Blundell, Alexander Pritzel, and Benjamin Van~Roy.
\newblock Deep exploration via bootstrapped dqn.
\newblock \emph{Advances in neural information processing systems}, 29, 2016{\natexlab{a}}.

\bibitem[Osband et~al.(2016{\natexlab{b}})Osband, Van~Roy, and Wen]{osband2016generalization}
Ian Osband, Benjamin Van~Roy, and Zheng Wen.
\newblock Generalization and exploration via randomized value functions.
\newblock In \emph{International Conference on Machine Learning}, pages 2377--2386. PMLR, 2016{\natexlab{b}}.

\bibitem[Kakade(2003)]{kakade2003sample}
Sham~Machandranath Kakade.
\newblock \emph{On the sample complexity of reinforcement learning}.
\newblock University of London, University College London (United Kingdom), 2003.

\bibitem[Kearns and Singh(2002)]{kearns2002near}
Michael Kearns and Satinder Singh.
\newblock Near-optimal reinforcement learning in polynomial time.
\newblock \emph{Machine learning}, 49\penalty0 (2):\penalty0 209--232, 2002.

\bibitem[Brafman and Tennenholtz(2002)]{brafman2002r}
Ronen~I Brafman and Moshe Tennenholtz.
\newblock R-max-a general polynomial time algorithm for near-optimal reinforcement learning.
\newblock \emph{Journal of Machine Learning Research}, 3\penalty0 (Oct):\penalty0 213--231, 2002.

\bibitem[O’Donoghue et~al.(2018)O’Donoghue, Osband, Munos, and Mnih]{o2018uncertainty}
Brendan O’Donoghue, Ian Osband, Remi Munos, and Volodymyr Mnih.
\newblock The uncertainty bellman equation and exploration.
\newblock In \emph{International conference on machine learning}, pages 3836--3845, 2018.

\bibitem[Bellemare et~al.(2016)Bellemare, Srinivasan, Ostrovski, Schaul, Saxton, and Munos]{bellemare2016unifying}
Marc Bellemare, Sriram Srinivasan, Georg Ostrovski, Tom Schaul, David Saxton, and Remi Munos.
\newblock Unifying count-based exploration and intrinsic motivation.
\newblock \emph{Advances in neural information processing systems}, 29, 2016.

\bibitem[Ostrovski et~al.(2017)Ostrovski, Bellemare, Oord, and Munos]{ostrovski2017count}
Georg Ostrovski, Marc~G Bellemare, A{\"a}ron Oord, and R{\'e}mi Munos.
\newblock Count-based exploration with neural density models.
\newblock In \emph{International conference on machine learning}, pages 2721--2730. PMLR, 2017.

\bibitem[Burda et~al.(2019)Burda, Edwards, Storkey, and Klimov]{burda2018exploration}
Yuri Burda, Harrison Edwards, Amos Storkey, and Oleg Klimov.
\newblock Exploration by random network distillation.
\newblock In \emph{International Conference on Learning Representations}, 2019.
\newblock URL \url{https://openreview.net/forum?id=H1lJJnR5Ym}.

\bibitem[Pathak et~al.(2017)Pathak, Agrawal, Efros, and Darrell]{pathak2017curiosity}
Deepak Pathak, Pulkit Agrawal, Alexei~A Efros, and Trevor Darrell.
\newblock Curiosity-driven exploration by self-supervised prediction.
\newblock In \emph{International conference on machine learning}, pages 2778--2787. PMLR, 2017.

\bibitem[Mannor et~al.(2007)Mannor, Simester, Sun, and Tsitsiklis]{mannor2007bias}
Shie Mannor, Duncan Simester, Peng Sun, and John~N Tsitsiklis.
\newblock Bias and variance approximation in value function estimates.
\newblock \emph{Management Science}, 53\penalty0 (2):\penalty0 308--322, 2007.

\bibitem[Lakshminarayanan et~al.(2017)Lakshminarayanan, Pritzel, and Blundell]{lakshminarayanan2017simple}
Balaji Lakshminarayanan, Alexander Pritzel, and Charles Blundell.
\newblock Simple and scalable predictive uncertainty estimation using deep ensembles.
\newblock \emph{Advances in neural information processing systems}, 30, 2017.

\bibitem[Gal and Ghahramani(2016)]{gal2016dropout}
Yarin Gal and Zoubin Ghahramani.
\newblock Dropout as a bayesian approximation: Representing model uncertainty in deep learning.
\newblock In \emph{international conference on machine learning}, pages 1050--1059. PMLR, 2016.

\bibitem[Lobel et~al.(2023)Lobel, Bagaria, and Konidaris]{Lobel2023Flipping}
Sam Lobel, Akhil Bagaria, and George Konidaris.
\newblock Flipping coins to estimate pseudocounts for exploration in reinforcement learning.
\newblock In \emph{International Conference on Machine Learning}, 2023.

\bibitem[Schulman et~al.(2017)Schulman, Wolski, Dhariwal, Radford, and Klimov]{schulman2017proximalpolicyoptimizationalgorithms}
John Schulman, Filip Wolski, Prafulla Dhariwal, Alec Radford, and Oleg Klimov.
\newblock Proximal policy optimization algorithms, 2017.
\newblock URL \url{https://arxiv.org/abs/1707.06347}.

\bibitem[O'Donoghue et~al.(2017)O'Donoghue, Osband, Munos, and Mnih]{ODonoghue2017TheUB}
Brendan O'Donoghue, Ian Osband, R{\'e}mi Munos, and Volodymyr Mnih.
\newblock The uncertainty bellman equation and exploration.
\newblock In \emph{International Conference on Machine Learning}, 2017.
\newblock URL \url{https://api.semanticscholar.org/CorpusID:6201691}.

\bibitem[Lattimore and Szepesv{\'a}ri(2020)]{lattimore2020bandit}
Tor Lattimore and Csaba Szepesv{\'a}ri.
\newblock \emph{Bandit algorithms}.
\newblock Cambridge University Press, 2020.

\bibitem[Auer et~al.(2008)Auer, Jaksch, and Ortner]{auer2008near}
Peter Auer, Thomas Jaksch, and Ronald Ortner.
\newblock Near-optimal regret bounds for reinforcement learning.
\newblock \emph{Advances in neural information processing systems}, 21, 2008.

\bibitem[Yue et~al.(2025)Yue, Yuan, Yu, Zuo, Zhu, Xu, Chen, Wang, Fan, Du, Wei, Yu, Liu, Liu, Liu, Lin, Lin, Ma, Zhang, Zhang, Zhang, Zhu, Zhang, Liu, Wang, Wu, and Yan]{yue2025vapoefficientreliablereinforcement}
Yu~Yue, Yufeng Yuan, Qiying Yu, Xiaochen Zuo, Ruofei Zhu, Wenyuan Xu, Jiaze Chen, Chengyi Wang, TianTian Fan, Zhengyin Du, Xiangpeng Wei, Xiangyu Yu, Gaohong Liu, Juncai Liu, Lingjun Liu, Haibin Lin, Zhiqi Lin, Bole Ma, Chi Zhang, Mofan Zhang, Wang Zhang, Hang Zhu, Ru~Zhang, Xin Liu, Mingxuan Wang, Yonghui Wu, and Lin Yan.
\newblock Vapo: Efficient and reliable reinforcement learning for advanced reasoning tasks, 2025.
\newblock URL \url{https://arxiv.org/abs/2504.05118}.

\bibitem[Liu et~al.(2025)Liu, Chen, Li, Qi, Pang, Du, Lee, and Lin]{liu2025understandingr1zeroliketrainingcritical}
Zichen Liu, Changyu Chen, Wenjun Li, Penghui Qi, Tianyu Pang, Chao Du, Wee~Sun Lee, and Min Lin.
\newblock Understanding r1-zero-like training: A critical perspective, 2025.
\newblock URL \url{https://arxiv.org/abs/2503.20783}.

\bibitem[Tang et~al.(2017{\natexlab{a}})Tang, Houthooft, Foote, Stooke, Xi~Chen, Duan, Schulman, DeTurck, and Abbeel]{NIPS2017_3a20f62a}
Haoran Tang, Rein Houthooft, Davis Foote, Adam Stooke, OpenAI Xi~Chen, Yan Duan, John Schulman, Filip DeTurck, and Pieter Abbeel.
\newblock \#exploration: A study of count-based exploration for deep reinforcement learning.
\newblock In I.~Guyon, U.~Von Luxburg, S.~Bengio, H.~Wallach, R.~Fergus, S.~Vishwanathan, and R.~Garnett, editors, \emph{Advances in Neural Information Processing Systems}, volume~30. Curran Associates, Inc., 2017{\natexlab{a}}.
\newblock URL \url{https://proceedings.neurips.cc/paper_files/paper/2017/file/3a20f62a0af1aa152670bab3c602feed-Paper.pdf}.

\bibitem[Wen et~al.(2024)Wen, Liao, Deng, Wang, Zhang, and Wen]{wen2024entropyregularizedtokenlevelpolicyoptimization}
Muning Wen, Junwei Liao, Cheng Deng, Jun Wang, Weinan Zhang, and Ying Wen.
\newblock Entropy-regularized token-level policy optimization for language agent reinforcement, 2024.
\newblock URL \url{https://arxiv.org/abs/2402.06700}.

\bibitem[Wang et~al.(2025)Wang, Yu, Gao, Zheng, Liu, Lu, Dang, Chen, Yang, Zhang, Liu, Yang, Zhao, Yue, Song, Yu, Huang, and Lin]{wang20258020rulehighentropyminority}
Shenzhi Wang, Le~Yu, Chang Gao, Chujie Zheng, Shixuan Liu, Rui Lu, Kai Dang, Xionghui Chen, Jianxin Yang, Zhenru Zhang, Yuqiong Liu, An~Yang, Andrew Zhao, Yang Yue, Shiji Song, Bowen Yu, Gao Huang, and Junyang Lin.
\newblock Beyond the 80/20 rule: High-entropy minority tokens drive effective reinforcement learning for llm reasoning, 2025.
\newblock URL \url{https://arxiv.org/abs/2506.01939}.

\bibitem[Cheng et~al.(2025)Cheng, Huang, Zhu, Dai, Zhao, Zhang, and Wei]{cheng2025reasoningexplorationentropyperspective}
Daixuan Cheng, Shaohan Huang, Xuekai Zhu, Bo~Dai, Wayne~Xin Zhao, Zhenliang Zhang, and Furu Wei.
\newblock Reasoning with exploration: An entropy perspective on reinforcement learning for llms, 2025.
\newblock URL \url{https://arxiv.org/abs/2506.14758}.

\bibitem[Zhang et~al.(2024)Zhang, Yu, Sharma, Yang, Wang, Hassan, and Wang]{zhang2024self}
Shenao Zhang, Donghan Yu, Hiteshi Sharma, Ziyi Yang, Shuohang Wang, Hany Hassan, and Zhaoran Wang.
\newblock Self-exploring language models: Active preference elicitation for online alignment.
\newblock \emph{arXiv preprint arXiv:2405.19332}, 2024.

\bibitem[Bai et~al.(2025)Bai, Zhan, Qiu, Zhang, Xu, and Li]{DPOCFN}
Chenjia Bai, Yang Zhan, Shuang Qiu, Qiaosheng Zhang, Kang Xu, and Xuelong Li.
\newblock Online preference alignment for language models via count-based exploration.
\newblock In \emph{International Conference on Learning Representations}, 2025.

\bibitem[Cen et~al.()Cen, Mei, Goshvadi, Dai, Yang, Yang, Schuurmans, Chi, and Dai]{cenvalue}
Shicong Cen, Jincheng Mei, Katayoon Goshvadi, Hanjun Dai, Tong Yang, Sherry Yang, Dale Schuurmans, Yuejie Chi, and Bo~Dai.
\newblock Value-incentivized preference optimization: A unified approach to online and offline rlhf.
\newblock In \emph{The Thirteenth International Conference on Learning Representations}.

\bibitem[Chen et~al.(2025)Chen, Chen, Sun, and Zhang]{chen2025avoidingmathbfexprmaxscalingrlhf}
Mingyu Chen, Yiding Chen, Wen Sun, and Xuezhou Zhang.
\newblock Avoiding $\mathbf{exp(R_{max})}$ scaling in rlhf through preference-based exploration, 2025.
\newblock URL \url{https://arxiv.org/abs/2502.00666}.

\bibitem[Gao et~al.(2025)Gao, Pan, Wang, Zhong, Lu, Cai, Jiang, and Zhao]{gao2025navigate}
Jingtong Gao, Ling Pan, Yejing Wang, Rui Zhong, Chi Lu, Qingpeng Cai, Peng Jiang, and Xiangyu Zhao.
\newblock Navigate the unknown: Enhancing llm reasoning with intrinsic motivation guided exploration.
\newblock \emph{arXiv preprint arXiv:2505.17621}, 2025.

\bibitem[Bai et~al.(2021)Bai, Wang, Han, Hao, Garg, Liu, and Wang]{pmlr-v139-bai21d}
Chenjia Bai, Lingxiao Wang, Lei Han, Jianye Hao, Animesh Garg, Peng Liu, and Zhaoran Wang.
\newblock Principled exploration via optimistic bootstrapping and backward induction.
\newblock In Marina Meila and Tong Zhang, editors, \emph{Proceedings of the 38th International Conference on Machine Learning}, volume 139 of \emph{Proceedings of Machine Learning Research}, pages 577--587. PMLR, 18--24 Jul 2021.
\newblock URL \url{https://proceedings.mlr.press/v139/bai21d.html}.

\bibitem[Tang et~al.(2017{\natexlab{b}})Tang, Houthooft, Foote, Stooke, Xi~Chen, Duan, Schulman, DeTurck, and Abbeel]{tang2017exploration}
Haoran Tang, Rein Houthooft, Davis Foote, Adam Stooke, OpenAI Xi~Chen, Yan Duan, John Schulman, Filip DeTurck, and Pieter Abbeel.
\newblock \# exploration: A study of count-based exploration for deep reinforcement learning.
\newblock \emph{Advances in neural information processing systems}, 30, 2017{\natexlab{b}}.

\bibitem[Rashid et~al.(2019)Rashid, Peng, Boehmer, and Whiteson]{rashid2019optimistic}
Tabish Rashid, Bei Peng, Wendelin Boehmer, and Shimon Whiteson.
\newblock Optimistic exploration even with a pessimistic initialisation.
\newblock In \emph{International Conference on Learning Representations}, 2019.

\bibitem[Yang et~al.(2024)Yang, Zhang, Hui, Gao, Yu, Li, Liu, Tu, Zhou, Lin, Lu, Xue, Lin, Liu, Ren, and Zhang]{yang2024qwen25mathtechnicalreportmathematical}
An~Yang, Beichen Zhang, Binyuan Hui, Bofei Gao, Bowen Yu, Chengpeng Li, Dayiheng Liu, Jianhong Tu, Jingren Zhou, Junyang Lin, Keming Lu, Mingfeng Xue, Runji Lin, Tianyu Liu, Xingzhang Ren, and Zhenru Zhang.
\newblock Qwen2.5-math technical report: Toward mathematical expert model via self-improvement, 2024.
\newblock URL \url{https://arxiv.org/abs/2409.12122}.

\bibitem[Hendrycks et~al.(2021)Hendrycks, Burns, Kadavath, Arora, Basart, Tang, Song, and Steinhardt]{hendrycks2021measuring}
Dan Hendrycks, Collin Burns, Saurav Kadavath, Akul Arora, Steven Basart, Eric Tang, Dawn Song, and Jacob Steinhardt.
\newblock Measuring mathematical problem solving with the {MATH} dataset.
\newblock In \emph{Thirty-fifth Conference on Neural Information Processing Systems Datasets and Benchmarks Track (Round 2)}, 2021.
\newblock URL \url{https://openreview.net/forum?id=7Bywt2mQsCe}.

\bibitem[He et~al.(2024)He, Luo, Bai, Hu, Thai, Shen, Hu, Han, Huang, Zhang, et~al.]{he2024olympiadbench}
Chaoqun He, Renjie Luo, Yuzhuo Bai, Shengding Hu, Zhen Thai, Junhao Shen, Jinyi Hu, Xu~Han, Yujie Huang, Yuxiang Zhang, et~al.
\newblock Olympiadbench: A challenging benchmark for promoting agi with olympiad-level bilingual multimodal scientific problems.
\newblock In \emph{Proceedings of the 62nd Annual Meeting of the Association for Computational Linguistics (Volume 1: Long Papers)}, pages 3828--3850, 2024.

\bibitem[Tang et~al.(2024)Tang, Zhang, Wang, and Wei]{tang2024mathscale}
Zhengyang Tang, Xingxing Zhang, Benyou Wang, and Furu Wei.
\newblock Mathscale: Scaling instruction tuning for mathematical reasoning.
\newblock In \emph{International Conference on Machine Learning}, pages 47885--47900. PMLR, 2024.

\bibitem[Lewkowycz et~al.(2022)Lewkowycz, Andreassen, Dohan, Dyer, Michalewski, Ramasesh, Slone, Anil, Schlag, Gutman-Solo, Wu, Neyshabur, Gur-Ari, and Misra]{lewkowycz2022solving}
Aitor Lewkowycz, Anders~Johan Andreassen, David Dohan, Ethan Dyer, Henryk Michalewski, Vinay~Venkatesh Ramasesh, Ambrose Slone, Cem Anil, Imanol Schlag, Theo Gutman-Solo, Yuhuai Wu, Behnam Neyshabur, Guy Gur-Ari, and Vedant Misra.
\newblock Solving quantitative reasoning problems with language models.
\newblock In Alice~H. Oh, Alekh Agarwal, Danielle Belgrave, and Kyunghyun Cho, editors, \emph{Advances in Neural Information Processing Systems}, 2022.
\newblock URL \url{https://openreview.net/forum?id=IFXTZERXdM7}.

\bibitem[Grattafiori et~al.(2024)Grattafiori, Dubey, Jauhri, Pandey, Kadian, Al-Dahle, Letman, Mathur, Schelten, Vaughan, and et~al]{grattafiori2024llama3herdmodels}
Aaron Grattafiori, Abhimanyu Dubey, Abhinav Jauhri, Abhinav Pandey, Abhishek Kadian, Ahmad Al-Dahle, Aiesha Letman, Akhil Mathur, Alan Schelten, Alex Vaughan, and et~al.
\newblock The llama 3 herd of models.
\newblock 2024.
\newblock URL \url{https://arxiv.org/abs/2407.21783}.

\bibitem[Li et~al.(2023)Li, Hui, Qu, Li, Yang, Li, Wang, Qin, Geng, Huo, Zhou, Ma, Li, Chang, Huang, Cheng, and Li]{li2023llm}
Jinyang Li, Binyuan Hui, Ge~Qu, Binhua Li, Jiaxi Yang, Bowen Li, Bailin Wang, Bowen Qin, Ruiying Geng, Nan Huo, Xuanhe Zhou, Chenhao Ma, Guoliang Li, Kevin C.~C. Chang, Fei Huang, Reynold Cheng, and Yongbin Li.
\newblock Can llm already serve as a database interface? a big bench for large-scale database grounded text-to-sqls, 2023.

\bibitem[Yu et~al.(2019)Yu, Zhang, Yang, Yasunaga, Wang, Li, Ma, Li, Yao, Roman, Zhang, and Radev]{yu2019spiderlargescalehumanlabeleddataset}
Tao Yu, Rui Zhang, Kai Yang, Michihiro Yasunaga, Dongxu Wang, Zifan Li, James Ma, Irene Li, Qingning Yao, Shanelle Roman, Zilin Zhang, and Dragomir Radev.
\newblock Spider: A large-scale human-labeled dataset for complex and cross-domain semantic parsing and text-to-sql task, 2019.
\newblock URL \url{https://arxiv.org/abs/1809.08887}.

\bibitem[Sheng et~al.(2025)Sheng, Zhang, Ye, Wu, Zhang, Zhang, Peng, Lin, and Wu]{Sheng_2025}
Guangming Sheng, Chi Zhang, Zilingfeng Ye, Xibin Wu, Wang Zhang, Ru~Zhang, Yanghua Peng, Haibin Lin, and Chuan Wu.
\newblock Hybridflow: A flexible and efficient rlhf framework.
\newblock In \emph{Proceedings of the Twentieth European Conference on Computer Systems}, EuroSys ’25, page 1279–1297. ACM, March 2025.
\newblock \doi{10.1145/3689031.3696075}.
\newblock URL \url{http://dx.doi.org/10.1145/3689031.3696075}.

\end{thebibliography}

\appendix
\section{Entropy Regularization as a Generalized $\epsilon$-Greedy Exploration}

We provide a mathematical derivation showing that entropy regularization corresponds to a softmax exploration strategy. This can be interpreted as a generalized form of $\epsilon$-greedy exploration that intelligently allocates exploration probability based on the relative quality of suboptimal actions.

\paragraph{Entropy-regularized policy improvement.}
Given a state $s$ and advantage estimates $A(s,a)$ for actions $a\in\mathcal{A}$, consider the entropy-regularized optimization:
\[
\pi^\star = \arg\max_{\pi(\cdot|s)} \;
\sum_{a}\pi(a|s)\,A(s,a) + \beta H(\pi(\cdot|s)),
\]
where $H(\pi) = -\sum_a \pi(a|s)\log \pi(a|s)$ and $\beta>0$ is the entropy coefficient.
The well-known solution is the Boltzmann/softmax distribution:
\[
\pi_\beta(a|s) = \frac{\exp(A(s,a)/\beta)}{\sum_{b \in \mathcal{A}} \exp(A(s,b)/\beta)}.
\]

\paragraph{Connection to $\epsilon$-greedy.}
Let $a^\star \in \arg\max_a A(s,a)$ and denote the advantage gaps $\Delta_a = A(s,a^\star)-A(s,a)\ge 0$. The probability of selecting the optimal action is
\[
p_\star = \pi_\beta(a^\star|s) =
\frac{1}{1+\sum_{a\ne a^\star}\exp(-\Delta_a/\beta)}.
\]
We can define a state- and value-dependent exploration probability $\epsilon_\beta(s) = 1-p_\star$. This allows us to decompose the policy as:
\[
\pi_\beta(\cdot|s) = (1-\epsilon_\beta(s))\,\delta_{a^\star}
+ \epsilon_\beta(s)\,q_\beta(\cdot|s),
\]
where $q_\beta(a|s) \propto \exp(-\Delta_a/\beta)$ is a probability distribution over the set of suboptimal actions.

This formulation reveals that softmax exploration is a generalized form of $\epsilon$-greedy. However, unlike the standard $\epsilon$-greedy rule, its exploration is not uniform. The distribution $q_\beta$ intelligently assigns higher probability to suboptimal actions that are closer to optimal (i.e., having a smaller advantage gap $\Delta_a$). Only under the strong and often unrealistic condition that all suboptimal actions are equally bad ($\Delta_a \approx \text{const.}$ for $a \ne a^\star$) does $q_\beta$ approach a uniform distribution, making the strategy resemble standard $\epsilon$-greedy. Thus, entropy regularization typically leads to a more efficient exploration strategy than its uniform counterpart.

\section{Proof of Proposition 1}

The proof follows the methodology presented in \citep{o2018uncertainty}. According to the definition of the conditional variance, we have:
\begin{align*}
\mathbb{V}_t[\hat{Q}^h(s,a)] &= \mathbf{E}_t \left[ (\hat{Q}^h(s,a) - \mathbf{E}_t[\hat{Q}^h(s,a)])^2 \right] \\
&= \mathbf{E}_t \left[ \left( \hat{r}^h(s) - \mathbf{E}_t[\hat{r}^h(s)] + \sum_{s',a'} \pi_{s'a'}^h P_{s'sa'}^h (\hat{Q}^{h+1}(s',a') - \mathbf{E}_t[\hat{Q}^{h+1}(s',a')]) \right)^2 \right] \\
&= \mathbf{E}_t \left[ (\hat{r}^h(s) - \mathbf{E}_t[\hat{r}^h(s)])^2 \right] \\
& \ \ \ \ \ + \mathbf{E}_t \left[ \left( \sum_{s',a'} \pi_{s'a'}^h P_{s'sa'}^h (\hat{Q}^{h+1}(s',a') - \mathbf{E}_t[\hat{Q}^{h+1}(s',a')]) \right)^2 \right]
\end{align*}
The second equality holds by expanding the square and assuming that the reward estimate $\hat{r}^h(s)$ and the next-step Q-value estimate $\hat{Q}^{h+1}(s',a')$ are conditionally independent, which makes their cross-product term zero.

Now, we focus on the second term. Noting that $\sum_{s',a'} \pi_{s'a'}^h P^{h}_{s'sa'} = 1$, this term represents a weighted sum. Since the function $f(x)=x^2$ is convex, we can apply \textbf{Jensen's inequality}.

For a convex function $f$, weights $w_i$ that sum to 1, and random variables $Z_i$, Jensen's inequality states:
\[
\mathbf{E} \left[ f\left( \sum_{i} w_i Z_i \right) \right] \leq \sum_{i} w_i \mathbf{E} \left[ f(Z_i) \right]
\]
Applying this to our expression gives:
\begin{align*}
\mathbf{E}_t & \left[ \left( \sum_{s',a'} \pi_{s'a'}^h P_{s'sa'}^h (\hat{Q}^{h+1}(s',a') - \mathbf{E}_t[\hat{Q}^{h+1}(s',a')]) \right)^2 \right] \\
&\leq \sum_{s',a'} \pi_{s'a'}^h P_{s'sa'}^h \, \mathbf{E}_t \left[ (\hat{Q}^{h+1}(s',a') - \mathbf{E}_t[\hat{Q}^{h+1}(s',a')])^2 \right] \\
&= \sum_{s',a'} \pi_{s'a'}^h P_{s'sa'}^h \, \mathbb{V}_t[\hat{Q}^{h+1}(s',a')]
\end{align*}

Combining the results, we arrive at the final inequality:
\[
\mathbb{V}_t[\hat{Q}^h(s,a)] \leq \mathbb{V}_t[\hat{r}^h(s)] + \sum_{s',a'} \pi_{s'a'}^h P_{s'sa'}^h \, \mathbb{V}_t[\hat{Q}^{h+1}(s',a')]
\]
This shows that the variance of the Q-value at step $h$ is bounded by the variance of the immediate reward plus the expected variance of the Q-value at the next step, $h+1$.

\newpage
\section{Detailed Training Configurations}
\label{details}
\subsection{Training Data and Reward Function}
\paragraph{Mathematical Reasoning}
For both our train dataset and test dataset, we use the following system prompt:
\begin{tcolorbox}[title=\textbf{System Prompt},colback=SeaGreen!10!CornflowerBlue!10,colframe=RoyalPurple!55!Aquamarine!100!]
Please reason step by step, and put your final answer within \textbackslash boxed\{\}.
\end{tcolorbox}

We use an outcome-based reward function that assigns +1 for correct final answers and -1 otherwise.

\paragraph{SQL Generation}
For both our train dataset and test dataset, we do not explicitly use any system prompt. We add the following contents at the beginning of the user prompt:
\begin{tcolorbox}[title=\textbf{Prompt},colback=SeaGreen!10!CornflowerBlue!10,colframe=RoyalPurple!55!Aquamarine!100!]
Task Overview:

You are a data science expert. Below, you are provided with a database schema and a natural language question. Your task is to understand the schema and generate a valid SQL query to answer the question.
\end{tcolorbox}

The outcome-based reward function is dense: $\text{final\_score} = \text{answer\_score}\ +\ \text{format\_score}$, where:
\begin{equation}
    \text{answer\_score} = \begin{cases}
        1.0, &  \text{if } \text{Result}(S) = \text{Result}(G) \\
        \min \left( \frac{\text{count}^2}{\left| \text{gold\_dict} \right| \times \left| \text{result\_dict} \right|}, 1.0 \right) \times 0.8 & \text{if } \text{Result}(S) \ne \text{Result}(G)
    \end{cases}
\end{equation}
Above, $S$ is the generated solution string (predicted SQL query), and $G$ is the ground truth query. $\text{Result}(Q)$ is the set of execution results returned by the database when executing the SQL query $Q$.

\subsection{CFN Training Configuration}
For CFN training, we first generate rollouts from the backbone model on the training dataset and use these responses to perform a preliminary training of the CFN. This process enables it to develop a basic understanding of which states are likely to occur more rarely. During the RL training phase, we initialize the exploration model with the parameters of the pretrained CFN to prevent the information it provides at the outset from misleading the policy model.

\paragraph{Mathmeatical Reasoning}
We use the hyperparameters in Table \ref{cfn-table1} for CFN training on mathematical reasoning tasks.
\begin{table}
  \caption{Our CFN training configurations on mathematical reasoning tasks.}
  \label{cfn-table1}
  \centering
  \begin{tabular}{llll}
    \toprule
    Hyperparameter     & Value  \\
    \midrule
    Optimizer & AdamW \\
    Learning rate in the pretraining process & 1e-3 \\
    Learning rate in the RL training process & 1e-4 \\
    Training batch size & $512 \times 8$ \\
    Mini-batch size & 8 \\
    \bottomrule
  \end{tabular}
\end{table}

\paragraph{SQL Generation}
We use the hyperparameters in Table \ref{cfn-table2} for CFN training on SQL generation tasks.
\begin{table}
  \caption{Our CFN training configurations on SQL generation tasks.}
  \label{cfn-table2}
  \centering
  \begin{tabular}{llll}
    \toprule
    Hyperparameter     & Value  \\
    \midrule
    Optimizer & AdamW \\
    Learning rate in the pretraining process & 3e-4 \\
    Learning rate in the RL training process & 1e-4 \\
    Training batch size & $128 \times 8$ \\
    Mini-batch size & 8 \\
    \bottomrule
  \end{tabular}
\end{table}

\subsection{RL Training Configuration}
\label{rl_config}

\paragraph{Mathmeatical Reasoning}
We use the hyperparameters in Table \ref{rl-table1} for RL training on mathematical reasoning tasks. Notably, during RL training, we applied a scheduled warm-up and cosine decay to the discount factor $\gamma$:

\begin{equation}
    \gamma=
\begin{cases}
\gamma_\text{max} \times \frac{t_{\text{iter}}}{T_{\text{start\_decay}}} & \text{if } t_{\text{iter}}\leq T_{\text{start\_decay}}\\
\beta \gamma_\text{max}+\left( 1-\beta \right) \times \gamma_\text{max} \times \frac{1}{2} \left(1 + \cos\left(\pi \times \frac{\min(t_{\text{iter}}, T_{\text{end\_decay}}) - T_{\text{start\_decay}}}{T_{\text{end\_decay}} - T_{\text{start\_decay}}}\right)\right) & \text{if } t_{\text{iter}} > T_{\text{start\_decay}}

\end{cases},
\end{equation}

where we set $T_{\text{start\_decay}}=10$, $T_{\text{end\_decay}}=200$, $\beta=0.1$. The same applies in the SQL generation task.

\begin{table}
  \caption{Our RL training configurations on mathematical reasoning tasks. }
  \label{rl-table1}
  \centering
  \begin{tabular}{llll}
    \toprule
    Hyperparameter     & Value  \\
    \midrule
    Optimizer & AdamW \\
    Policy learning rate & 1e-6 \\
    Training batch size & 512 \\
    Samples per prompt & 8 \\
    Mini-batch size & 32 \\
    Max prompt length & 1024 \\
    Max response length & 3072 \\
    Rollout temperature & 1.0 \\
    Top $p\%$ in step 1 of bonus filtering & $30\%$ \\
    Maximum $\gamma$ in Equation \ref{oldtonew} & 0.4 (GRPO) \& 0.2 (DAPO) \\
    $\alpha$ in Equation \ref{oldtonew} & 0.5 \\
    \bottomrule
  \end{tabular}
\end{table}

In addition to focusing on mean@\emph{k}, we also place considerable emphasis on pass@\emph{k}. However, we observe that as vanilla GRPO training progresses, increases in mean@\emph{k} are generally accompanied by sharp decreases in pass@\emph{k}, which is also presented in Appendix \ref{scale}. Therefore, to ensure comparability across both types of metrics, we train each experiment for 120 steps on vanilla GRPO. For DAPO, we train each experiment for 160 steps (including data sampling and filtering).

\paragraph{SQL Generation}
We use the hyperparameters in Table \ref{rl-table2} for RL training on SQL generation tasks. 

\begin{table}
  \caption{Our RL training configurations on SQL generation tasks.}
  \label{rl-table2}
  \centering
  \begin{tabular}{llll}
    \toprule
    Hyperparameter     & Value  \\
    \midrule
    Optimizer & AdamW \\
    Policy learning rate & 1e-6 \\
    Training batch size & 128 \\
    Samples per prompt & 8 \\
    Mini-batch size & 64 \\
    Max prompt length & 8192 \\
    Max response length & 4096 \\
    Rollout temperature & 1.0 \\
    Top $p\%$ in step 1 of bonus filtering & $20\%$ \\
    Maximum $\gamma$ in Equation \ref{oldtonew} & 0.1 \\
    $\alpha$ in Equation \ref{oldtonew} & 0.5 \\
    \bottomrule
  \end{tabular}
\end{table}

We train each experiment for 160 steps on vanilla GRPO, and 240 steps on DAPO.

\section{Inference Configurations}
\paragraph{Mathmeatical Reasoning}
We use a rollout temperature of $0.6$, top-$p$ sampling with $p=0.95$, and a maximum response length of 4096 tokens. We adopt $k=256$ for the small but challenging 
AIME2024/2025 datasets (30 problems each), $k=16$ for Minerva, MATH500, and OlympiadBench, and $k=8$ for College Math, balancing computational cost and difficulty.

\paragraph{SQL Generation}
We use a top-$p$ sampling with $p=0.95$, and a maximum response length of 4096 tokens. We use a rollout temperature of $0.0$ for greedy sampling, and a rollout temperature of $1.0$ to evaluate pass@\emph{k}.

\section{Additional Experimental Results}
\label{add_results}

\subsection{Scaling Experiments}
\label{scale}
As a comparison, we scaled the vanilla GRPO baseline experiment, continuing to train the GRPO baseline to the $260 th$ training step. We observe that, although extended training increases mean@\emph{k}, it substantially degrades pass@\emph{k}, consistent with the limitations of GRPO discussed in Appendix \ref{rl_config}. By contrast, our MERCI algorithm realizes its exploratory potential earlier, rapidly identifying good solutions and exhibiting improvements in pass@\emph{k} as well.

\begin{table*}[t]
\centering
\caption{Results of scaling experiments on vanilla GRPO and mathematical reasoning benchmarks. *GRPO-scaling is the vanilla GRPO checkpoint at step 260.}

% ---------- Pass@k ----------
\begin{subtable}{\textwidth}
\centering
\caption{pass@\emph{k} results}
\small
\setlength{\tabcolsep}{4.5pt}
\renewcommand{\arraystretch}{1.1}
\begin{tabular}{lcccccccc}
\toprule
& \textbf{AIME25} & \textbf{AIME24} & \textbf{Minerva} & \textbf{MATH500} & \textbf{OlympiadBench} & \textbf{College} & \textbf{Avg.} \\
& \textit{pass@256} & \textit{pass@256} & \textit{pass@16} & \textit{pass@16} & \textit{pass@16} & \textit{pass@8} & \\
\midrule
GRPO    &  50.0 & 76.7 & 64.0 & 91.8 & 59.7 & 49.2 & 65.8 \\
GRPO-scaling* & 46.7 & 70.0 & 59.6 & 88.8 & 56.4 & 48.1 & 61.6\\
GRPO + MERCI     &  60.0 & 80.0 & 63.2 & 91.4 & 60.9 & 48.9 & \best{67.4} \\
\bottomrule
\end{tabular}
\end{subtable}

\vspace{0.8em}

% ---------- mean@k ----------
\begin{subtable}{\textwidth}
\centering
\caption{mean@\emph{k} results}
\small
\setlength{\tabcolsep}{4.5pt}
\renewcommand{\arraystretch}{1.1}
\begin{tabular}{lcccccccc}
\toprule
& \textbf{AIME25} & \textbf{AIME24} & \textbf{Minerva} & \textbf{MATH500} & \textbf{OlympiadBench} & \textbf{College} & \textbf{Avg.} \\
& \textit{mean@256} & \textit{mean@256}  & \textit{mean@16} & \textit{mean@16} & \textit{mean@16} & \textit{mean@8} & \\
\midrule
GRPO         & 11.2 & 28.7 & 41.8 & 79.0 & 40.3 & 42.0 & 40.5 \\
GRPO-scaling & 12.7 & 28.3 & 42.8 & 78.7 & 40.8 & 42.7 & 41.0 \\
GRPO + MERCI   & 13.4 & 29.6 & 44.1 & 80.7 & 42.6 & 42.9 & \best{42.2} \\
\bottomrule
\end{tabular}
\end{subtable}
\end{table*}

\subsection{Training Dynamics}
\begin{figure}
  \centering
  \includegraphics[width=1.0\linewidth]{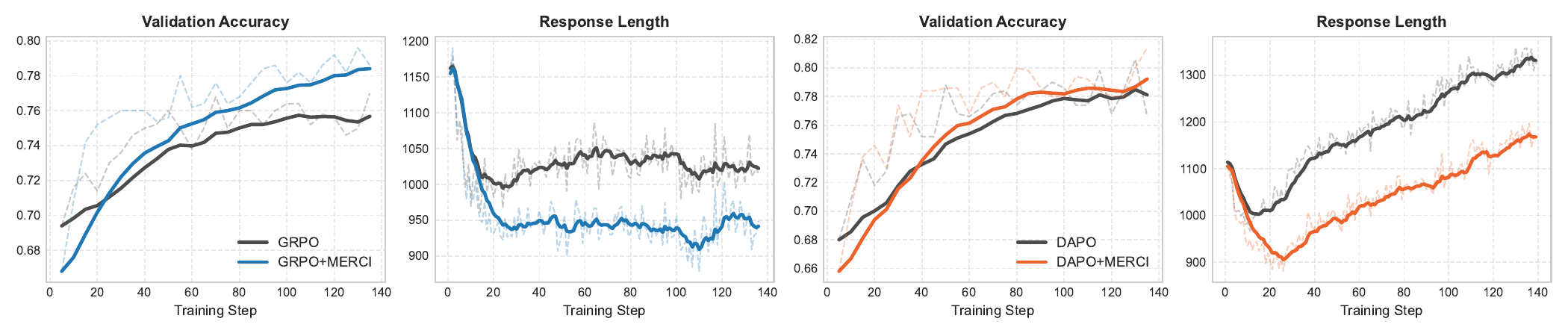}
  \caption{Validation (i.e., MATH500) accuracy and response length during training.}
  \label{training}
\end{figure}

We report the training dynamics of both validation accuracy and response length on mathematical reasoning tasks in Figure \ref{training}.

\subsection{Ablation Studies}
\label{ablation}
To verify the effectiveness of the modules in our method, we conducted ablation studies on the mathematical reasoning task and vanilla GRPO, and present the results in Table \ref{ablation_results}.

\begin{table*}[t]
\centering
\caption{Ablation studies on vanilla GRPO and mathematical reasoning benchmarks. *\textit{p \& s filtering} is a reduced-form representation of percentile and spatial coherence filtering. **The difference between \textit{cumulative std} and \textit{cumulative variance} (i.e, our main results) has been introduced in Paragraph 1 of Section \ref{estimation}. $\dag$ The setting of token integration expresses that, rather than computing uncertainty at the trajectory level for entire trajectories, we directly add an uncertainty estimate to each token-level advantage.}
\label{ablation_results}

% ---------- Pass@k ----------
\begin{subtable}{\textwidth}
\centering
\caption{pass@\emph{k} results}
\scriptsize
\setlength{\tabcolsep}{4.5pt}
\renewcommand{\arraystretch}{1.1}
\begin{tabular}{lcccccccc}
\toprule
& \textbf{AIME25} & \textbf{AIME24} & \textbf{Minerva} & \textbf{MATH500} & \textbf{OlympiadBench} & \textbf{College} & \textbf{Avg.} \\
& \textit{pass@256} & \textit{pass@256} & \textit{pass@16} & \textit{pass@16} & \textit{pass@16} & \textit{pass@8} & \\
\midrule
GRPO    &  50.0 & 76.7 & 64.0 & 91.8 & 59.7 & 49.2 & 65.8 \\
\midrule
GRPO + MERCI w/o \textit{p \& s filtering}* & 56.7 & 73.3 & 65.8 & 90.0 & 60.0 & 47.9 & 65.6 \\
GRPO + MERCI w/o noise filtering & 50.0 & 73.3 & 61.8 & 89.2 & 59.7 & 48.9 & 63.8 \\
\midrule
GRPO + MERCI w/ \textit{cumulative std}** &  56.7 &  76.7 & 65.1 & 90.6 & 60.2 & 48.1 & 66.2 \\
GRPO + MERCI w/ token integration$\dag$ & 46.7 & 70.0 & 62.5 & 91.6 & 59.3 & 49.1 & 63.2 \\
\midrule
GRPO + MERCI     &  60.0 & 80.0 & 63.2 & 91.4 & 60.9 & 48.9 & \best{67.4} \\
\bottomrule
\end{tabular}
\end{subtable}

\vspace{0.8em}

% ---------- mean@k ----------
\begin{subtable}{\textwidth}
\centering
\caption{mean@\emph{k} results}
\scriptsize
\setlength{\tabcolsep}{4.5pt}
\renewcommand{\arraystretch}{1.1}
\begin{tabular}{lcccccccc}
\toprule
& \textbf{AIME25} & \textbf{AIME24} & \textbf{Minerva} & \textbf{MATH500} & \textbf{OlympiadBench} & \textbf{College} & \textbf{Avg.} \\
& \textit{mean@256} & \textit{mean@256}  & \textit{mean@16} & \textit{mean@16} & \textit{mean@16} & \textit{mean@8} & \\
\midrule
GRPO         & 11.2 & 28.7 & 41.8 & 79.0 & 40.3 & 42.0 & 40.5 \\
\midrule
GRPO + MERCI w/o \textit{p \& s filtering} & 11.7 & 28.1 & 44.9 & 79.9 & 39.9 & 42.6 & 41.2 \\
GRPO + MERCI w/o noise filtering & 9.8 & 25.8 & 40.6 & 77.3 & 37.9 & 41.9 & 38.9 \\
\midrule
GRPO + MERCI w/ \textit{cumulative std} &  14.2 & 29.1 & 43.8 & 79.8 & 41.2 & 43.0 & 41.9 \\
GRPO + MERCI w/ token integration & 12.0 & 23.7 & 40.2 & 77.5 & 39.9 & 42.4 & 39.3 \\
\midrule
GRPO + MERCI   & 13.4 & 29.6 & 44.1 & 80.7 & 42.6 & 42.9 & \best{42.2} \\
\bottomrule
\end{tabular}
\end{subtable}
\end{table*}

As evidenced by the preceding experimental results, both noise filtering and normalized trajectory-level uncertainty estimation are critical to our method; without them, training can become unstable and may even underperform the baseline algorithm. Furthermore, percentile and spatial coherence filtering direct attention to uncertainty at key positions, while the variance-accumulation method yields more accurate estimates, thereby further improving our algorithm’s performance.

\subsection{Case Study}
\label{case}
We analyzed a case on AIME2024 to examine the effect of incorporating our method, and the results are as follows.

\begin{tcolorbox}[title=\textbf{Case Study},colback=SeaGreen!10!CornflowerBlue!10,colframe=RoyalPurple!55!Aquamarine!100!, breakable]
\textbf{Question:} Let $x,y$ and $z$ be positive real numbers that satisfy the following system of equations: 

$log_2\left({x \over yz}\right) = {1 \over 2}$ \ \ \ \ 
$log_2\left({y \over xz}\right) = {1 \over 3}$ \ \ \ \ 
$log_2\left({z \over xy}\right) = {1 \over 4}$

Then the value of $\left| log_2(x^4y^3z^2) \right|$ is $\tfrac{m}{n}$ where $m$ and $n$ are relatively prime positive integers. Find $m+n$.

\tcblower
\textbf{Response (DAPO):} Let's solve the given system of equations step by step. The system of equations provided is:

\[\log_2\left(\frac{x}{yz}\right) = \frac{1}{2}\]\[\log_2\left(\frac{y}{xz}\right) = \frac{1}{3}\]\[\log_2\left(\frac{z}{xy}\right) = \frac{1}{4}\]

First, we can rewrite each logarithmic equation in exponential form:1. \(\log_2\left(\frac{x}{yz}\right) = \frac{1}{2}\) implies \(\frac{x}{yz} = 2^{\frac{1}{2}} = \sqrt{2}\), so \(x = \sqrt{2} \cdot yz\).2. \(\log_2\left(\frac{y}{xz}\right) = \frac{1}{3}\) implies \(\frac{y}{xz} = 2^{\frac{1}{3}}\), so \(y = 2^{\frac{1}{3}} \cdot xz\).3. \(\log_2\left(\frac{z}{xy}\right) = \frac{1}{4}\) implies \(\frac{z}{xy} = 2^{\frac{1}{4}}\), so \(z = 2^{\frac{1}{4}} \cdot xy\).

Now, we can substitute the expressions for \(x\), \(y\), and \(z\) into each other to find the values of \(x\), \(y\), and \(z\):

From the first equation: \(x = \sqrt{2} \cdot yz\)

From the second equation: \(y = 2^{\frac{1}{3}} \cdot xz\)

From the third equation: \(z = 2^{\frac{1}{4}} \cdot xy\)

By multiplying these three equations together:\[x \cdot y \cdot z = (\sqrt{2} \cdot yz) \cdot (2^{\frac{1}{3}} \cdot xz) \cdot (2^{\frac{1}{4}} \cdot xy)\]\[x \cdot y \cdot z = \sqrt{2} \cdot 2^{\frac{1}{3}} \cdot 2^{\frac{1}{4}} \cdot (yz \cdot xz \cdot xy)\]\[x \cdot y \cdot z = \sqrt{2} \cdot 2^{\frac{1}{3}} \cdot 2^{\frac{1}{4}} \cdot (x^2 y^2 z^2)\]\[x \cdot y \cdot z = 2^{\frac{1}{2}} \cdot 2^{\frac{1}{3}} \cdot 2^{\frac{1}{4}} \cdot (x^2 y^2 z^2)\]\[x \cdot y \cdot z = 2^{\frac{1}{2} + \frac{1}{3} + \frac{1}{4}} \cdot (x^2 y^2 z^2)\]\[x \cdot y \cdot z = 2^{\frac{6}{12} + \frac{4}{12} + \frac{3}{12}} \cdot (x^2 y^2 z^2)\]\[x \cdot y \cdot z = 2^{\frac{13}{12}} \cdot (x^2 y^2 z^2)\]\[x \cdot y \cdot z = 2^{\frac{13}{12}} \cdot (x^2 y^2 z^2)\]\[1 = 2^{\frac{13}{12}} \cdot (xyz)\]\[1 = 2^{\frac{13}{12}} \cdot (xyz)\]\[xyz = 2^{-\frac{13}{12}}\]

Now we can use \(x = \sqrt{2} \cdot yz\), \(y = 2^{\frac{1}{3}} \cdot xz\), and \(z = 2^{\frac{1}{4}} \cdot xy\) to find the individual values of \(x\), \(y\), and \(z\):From \(x = \sqrt{2} \cdot yz\):\[ x = \sqrt{2} \cdot yz \]\[ x = \sqrt{2} \cdot \left(2^{-\frac{13}{12} \cdot \frac{1}{x}}\right) \]

Let’s find the exponents of \(x\), \(y\), and \(z\) in the expression \(x = \sqrt{2} \cdot yz\):\[x = \sqrt{2} \cdot yz\]\[x = 2^{\frac{1}{2}} \cdot yz\]\[x^2 = 2 \cdot y^2 z^2\]We know \(xyz = 2^{-\frac{13}{12}}\), so:\[x = 2^{\frac{1}{2}} \cdot \frac{2^{-\frac{13}{12}}}{x}\]\[x^2 = 2^{\frac{1}{2}} \cdot 2^{-\frac{13}{12}}\]\[x^2 = 2^{\frac{1}{2} - \frac{13}{12}}\]\[x^2 = 2^{\frac{6}{12} - \frac{13}{12}}\]\[x^2 = 2^{-\frac{7}{12}}\]\[x = 2^{-\frac{7}{24}}\]

Similarly:\[y = 2^{\frac{1}{3}} \cdot xz\]\[y = 2^{\frac{1}{3}} \cdot \frac{2^{-\frac{13}{12}}}{y}\]\[y^2 = 2^{\frac{1}{3}} \cdot 2^{-\frac{13}{12}}\]\[y^2 = 2^{\frac{1}{3} - \frac{13}{12}}\]\[y^2 = 2^{\frac{4}{12} - \frac{13}{12}}\]\[y^2 = 2^{-\frac{9}{12}}\]\[y^2 = 2^{-\frac{3}{4}}\]\[y = 2^{-\frac{3}{8}}\]Similarly:\[z = 2^{\frac{1}{4}} \cdot xy\]\[z = 2^{\frac{1}{4}} \cdot 2^{-\frac{7}{24}} \cdot 2^{-\frac{3}{8}}\]\[z = 2^{\frac{1}{4}} \cdot 2^{-\frac{7}{24} - \frac{3}{8}}\]\[z = 2^{\frac{1}{4}} \cdot 2^{-\frac{7}{24} - \frac{9}{24}}\]\[z = 2^{\frac{1}{4}} \cdot 2^{-\frac{16}{24}}\]\[z = 2^{\frac{1}{4}} \cdot 2^{-\frac{2}{3}}\]\[z = 2^{\frac{3}{12} - \frac{8}{12}}\]\[z = 2^{-\frac{5}{12}}\]

Now we need to find \( \left|\log_2(x^4 y^3 z^2)\right| \):\[x^4 y^3 z^2 = \left(2^{-\frac{7}{24}}\right)^4 \cdot \left(2^{-\frac{3}{8}}\right)^3 \cdot \left(2^{-\frac{5}{12}}\right)^2\]\[x^4 y^3 z^2 = 2^{-\frac{7}{6}} \cdot 2^{-\frac{9}{8}} \cdot 2^{-\frac{5}{6}}\]\[x^4 y^3 z^2 = 2^{-\frac{7}{6} - \frac{9}{8} - \frac{5}{6}}\]\[x^4 y^3 z^2 = 2^{-\frac{56}{48} - \frac{54}{48} - \frac{40}{48}}\]\[x^4 y^3 z^2 = 2^{-\frac{150}{48}} = 2^{-\frac{25}{8}}\]

Now we find the logarithm:\[\left|\log_2 (x^4 y^3 z^2)\right| = \left| \log_2 \left( 2^{-\frac{25}{8}} \right) \right| = \left| -\frac{25}{8} \right| = \frac{25}{8}\]

Thus, \(m = 25\) and \(n = 8\), so \(m + n = 33\).

Let's confirm with Python:

\begin{pycodebox}
# Calculating the valueimport math

# Values found

x_val = 2**(-7/24)y_val = 2**(-3/8)

z_val = 2**(-5/12)

# Calculate the expression x^4 * y^3 * z^2

expression_value = x_val**4 * y_val**3 * z_val**2

# Calculate log2 of the expression value

log_value = math.log2(expression_value)

# Absolute value

abs_log_value = abs(log_value)

print(abs_log_value)

\end{pycodebox}

\begin{outputbox}
$3.125$
\end{outputbox}

The calculation confirms that \( \left|\log_2(x^4 y^3 z^2)\right| = \frac{25}{8} = 3.125\), which matches our derived value. Thus, \( m = 25 \) and \( n = 8 \), so \( m + n = 33 \).

Therefore, the answer is:\[\boxed{33}\]

\textbf{Response (DAPO+MERCI):} To solve the given system of equations, we start by rewriting the logarithmic equations in exponential form:

1. \(\log_2\left(\frac{x}{yz}\right) = \frac{1}{2}\)

2. \(\log_2\left(\frac{y}{xz}\right) = \frac{1}{3}\)

3. \(\log_2\left(\frac{z}{xy}\right) = \frac{1}{4}\)

Using the definition of logarithms, we can convert these into the following equations:

1. \(\frac{x}{yz} = 2^{\frac{1}{2}} = \sqrt{2}\)

2. \(\frac{y}{xz} = 2^{\frac{1}{3}} = \sqrt[3]{2}\)

3. \(\frac{z}{xy} = 2^{\frac{1}{4}} = \sqrt[4]{2}\)

From these equations, we get:\[ x = \sqrt{2} \cdot yz \tag{1} \]\[ y = \sqrt[3]{2} \cdot xz \tag{2} \]\[ z = \sqrt[4]{2} \cdot xy \tag{3} \]

Now we substitute equation (1) into equation (2):\[ y = \sqrt[3]{2} \cdot (\sqrt{2} \cdot yz)z = \sqrt[3]{2} \cdot \sqrt{2} \cdot yz^2 = 2^{\frac{1}{3} + \frac{1}{2}} \cdot yz^2 = 2^{\frac{5}{6}} \cdot yz^2 \]Dividing both sides by \(y\), we get:\[ 1 = 2^{\frac{5}{6}} \cdot z^2 \]So,\[ z^2 = 2^{-\frac{5}{6}} \]\[ z = 2^{-\frac{5}{12}} \tag{4} \]

Now we substitute equation (1) into equation (3):\[ z = \sqrt[4]{2} \cdot (\sqrt{2} \cdot yz) y = \sqrt[4]{2} \cdot \sqrt{2} \cdot y^2 z = 2^{\frac{1}{4} + \frac{1}{2}} \cdot y^2 z = 2^{\frac{3}{4}} \cdot y^2 z \]Dividing both sides by \(z\), we get:\[ 1 = 2^{\frac{3}{4}} \cdot y^2 \]So,\[ y^2 = 2^{-\frac{3}{4}} \]\[ y = 2^{-\frac{3}{8}} \tag{5} \]

Now we substitute equation (2) into equation (3):\[ z = \sqrt[4]{2} \cdot (\sqrt[3]{2} \cdot xz) x = \sqrt[4]{2} \cdot \sqrt[3]{2} \cdot x^2 z = 2^{\frac{1}{4} + \frac{1}{3}} \cdot x^2 z = 2^{\frac{7}{12}} \cdot x^2 z \]Dividing both sides by \(z\), we get:\[ 1 = 2^{\frac{7}{12}} \cdot x^2 \]So,\[ x^2 = 2^{-\frac{7}{12}} \]\[ x = 2^{-\frac{7}{24}} \tag{6} \]

Now that we have the values:\[ x = 2^{-\frac{7}{24}}, \quad y = 2^{-\frac{3}{8}}, \quad z = 2^{-\frac{5}{12}} \]

We need to find the value of \(\left| \log_2(x^4 y^3 z^2) \right|\).

First, we calculate \(x^4 y^3 z^2\):\[ x^4 = \left(2^{-\frac{7}{24}}\right)^4 = 2^{-\frac{7}{6}} \]\[ y^3 = \left(2^{-\frac{3}{8}}\right)^3 = 2^{-\frac{9}{8}} \]\[ z^2 = \left(2^{-\frac{5}{12}}\right)^2 = 2^{-\frac{5}{6}} \]

So,\[ x^4 y^3 z^2 = 2^{-\frac{7}{6}} \cdot 2^{-\frac{9}{8}} \cdot 2^{-\frac{5}{6}} = 2^{-\frac{7}{6} - \frac{9}{8} - \frac{5}{6}} = 2^{-\left(\frac{7}{6} + \frac{9}{8} + \frac{5}{6}\right)} \]

To add the exponents, we need a common denominator:\[ \frac{7}{6} + \frac{9}{8} + \frac{5}{6} = \frac{56}{48} + \frac{54}{48} + \frac{40}{48} = \frac{56 + 54 + 40}{48} = \frac{150}{48} = \frac{25}{8} \]

So,\[ x^4 y^3 z^2 = 2^{-\frac{25}{8}} \]

Then,\[ \log_2(x^4 y^3 z^2) = \log_2\left(2^{-\frac{25}{8}}\right) = -\frac{25}{8} \]

Thus,\[ \left| \log_2(x^4 y^3 z^2) \right| = \left| -\frac{25}{8} \right| = \frac{25}{8} \]

The fraction \(\frac{25}{8}\) is already in its simplest form, so \(m = 25\) and \(n = 8\). Therefore, \(m + n = 25 + 8 = 33\).

The final answer is:\[\boxed{33}\]

\end{tcolorbox}

\end{document}